\documentclass[10pt,journal,compsoc]{IEEEtran}

\usepackage{soul,framed}

\usepackage{colortbl}

\usepackage[dvipsnames]{xcolor}

\usepackage[pdftex]{graphicx}
\graphicspath{{../pdf/}{../jpeg/}}
\DeclareGraphicsExtensions{.pdf,.jpeg,.png}

\usepackage[cmex10]{amsmath}
\usepackage{array}
\usepackage{mdwmath}
\usepackage{mdwtab}
\usepackage{eqparbox}
\usepackage{url}
\usepackage{graphicx}
\usepackage{amsmath}
\usepackage{amsfonts}
\usepackage{algorithm}
\usepackage{algpseudocode}
\usepackage{multirow}
\usepackage{footnote}
\usepackage{bm}
\usepackage{makecell}
\usepackage{pdflscape}
\usepackage{afterpage}
\usepackage{bbding}
\usepackage{lscape}
\usepackage{booktabs}
\usepackage{tikz}
\usepackage{ragged2e}
\usepackage{amssymb}
\usepackage{pifont}
\newcommand{\xmark}{\ding{55}}%

\usepackage{pgfplots}
\usepackage{subcaption}

\usepackage[square,sort,comma,numbers]{natbib} 
\usepackage[top=0.4in,bottom=0.4in,left=0.4in,textwidth=7.7in]{geometry}

\usepackage{enumitem}

\usepackage[pagebackref=false,breaklinks=true,linkcolor=red,anchorcolor=black, citecolor=black,colorlinks,bookmarks=true]{hyperref}
\usepackage[capitalize]{cleveref}
\crefname{section}{Sec.}{Secs.}
\Crefname{section}{Section}{Sections}
\Crefname{table}{Table}{Tables}
\crefname{table}{Tab.}{Tabs.}

\newcommand{\etal}{\textit{et al}. }

\newcommand{\ie}{\textit{i}.\textit{e}., }
\newcommand{\eg}{\textit{e}.\textit{g}., }
\newcommand{\vs}{\textit{vs.}\;}
\usepackage{amsfonts}
\usepackage{pifont}

\definecolor{mygreen}{RGB}{0, 102, 0}
\definecolor{myred}{RGB}{153, 0, 0}

\pgfplotsset{
    layers/my layer set/.define layer set={
        background,
        main,
        foreground
    }{ },
    set layers=my layer set,
}
\pgfplotsset{compat=1.12}

\begin{document}

\title{Rapid Salient Object Detection with Difference Convolutional Neural Networks}
 
\author{\IEEEauthorblockN{Zhuo Su\thanks{Zhuo Su is with the College of Computer Science, Nankai University, Tianjin, China, and also with the Center for Machine Vision and Signal Analysis (CMVS),
University of Oulu, Finland (email: zuike2013@outlook.com). The work was partially completed while Zhuo Su was interning at Intel Labs, Germany. Matthias Müller and Diana Wofk are with Intel Labs, Germany. Jiehua Zhang and  Matti Pietik\"{a}inen are with CMVS, University of Oulu, Finland. Ming-Ming Cheng is with the College of Computer Science, Nankai University, Tianjin, China. Li Liu is with the College of Electronic Science and Technology, NUDT, China. Corresponding authors: Li Liu (dreamliu2010@gmail.com) and Jiehua Zhang (jiehua.zhang@oulu.fi).},
Li Liu,
Matthias Müller,
Jiehua Zhang,
Diana Wofk,
Ming-Ming Cheng,
Matti Pietik\"{a}inen\thanks{This work was partially supported by the National Key Research and Development Program of China No. 2021YFB3100800, the Academy of Finland under grant 331883, and the National Natural Science Foundation of China under Grant 62376283. Code will be available at \url{https://github.com/hellozhuo/stdnet.git}.}}
}

\markboth{Accepted in IEEE TRANSACTIONS ON PATTERN ANALYSIS AND MACHINE INTELLIGENCE}
{Su \MakeLowercase{\textit{et al.}}: }

\IEEEtitleabstractindextext{%
\begin{abstract}
\justifying
This paper addresses the challenge of deploying salient object detection (SOD) on resource-constrained devices with real-time performance. While recent advances in deep neural networks have improved SOD, existing top-leading models are computationally expensive. We propose an efficient network design that combines traditional wisdom on SOD and the representation power of modern CNNs. Like biologically-inspired classical SOD methods relying on computing contrast cues to determine saliency of image regions, our model leverages Pixel Difference Convolutions (PDCs) to encode the feature contrasts. Differently, PDCs are incorporated in a CNN architecture so that the valuable contrast cues are extracted from rich feature maps. For efficiency, we introduce a difference convolution reparameterization (DCR) strategy that embeds PDCs into standard convolutions, eliminating computation and parameters at inference. Additionally, we introduce SpatioTemporal Difference Convolution (STDC) for video SOD, enhancing the standard 3D convolution with spatiotemporal contrast capture. Our models, SDNet for image SOD and STDNet for video SOD, achieve significant improvements in efficiency-accuracy trade-offs. On a Jetson Orin device, our models with $<$ 1M parameters operate at 46 FPS and 150 FPS on streamed images and videos, surpassing the second-best lightweight models in our experiments by more than $2\times$ and $3\times$ in speed with superior accuracy.
\end{abstract}

\begin{IEEEkeywords}
Real-time models, Image and video salient object detection, Convolutional neural networks, Pixel difference convolution
\end{IEEEkeywords}}

\twocolumn[
\begin{center}
    \textit{This work has been accepted for publication in IEEE Transactions on Pattern Analysis and Machine Intelligence (TPAMI). ©2025 IEEE.\\
    Personal use of this material is permitted. However, permission to reprint/republish this material for advertising or promotional purposes or for creating new collective works for resale or redistribution must be obtained from the IEEE.\\
    The work was initially done in March, 2023.}
\end{center}
\vspace{1em}
]

\maketitle
\IEEEdisplaynontitleabstractindextext
\IEEEpeerreviewmaketitle

\IEEEraisesectionheading{
\section{Introduction}}
\label{sec:intro}

Salient Object Detection (SOD) aims to segment the most visually distinctive (\ie salient) regions within an image or a video frame similar to the Human Visual System (HVS)~\cite{itti1998itti,liu2010learning}, and is formulated as a binary segmentation task in computer vision. Clearly,
humans are able to detect visually salient objects effortlessly and rapidly (\ie pre-attentive vision)~\cite{appelbaum2009attentive}; these filtered regions are then perceived with finer details to get richer high-level information (\ie attentive vision)~\cite{borji2019sodsurvey}. Likewise, SOD is a pre-attentive vision task that can benefit various more complex down-stream applications, including object recognition and detection~\cite{video_object_detection,rutishauser2004app-objectdetection,ren2013app-objectrecognition, chao-xiao}, semantic segmentation~\cite{app_segmentation,gao2022app-segmentation,ji2021full}, object tracking~\cite{hong2015app-tracking,wu2014weightedtracking}, image retrieval~\cite{he2012app-imageretrieval,cheng2014app-imageretrieval2}, image and video compression~\cite{patel2021imagecompression,itti2004videocompression1}, visual enhancement~\cite{app-enhancement,app-enhancement2}, image editing and augmentation~\cite{miangoleh2023realistic,uddin2020saliencymix}, and visual saliency modeling~\cite{borji2012quantitative}. It has also been used in other fields like computer graphics~\cite{lee2005mesh}, medical image analysis~\cite{ning2021smu} and remote sensing image analysis~\cite{li2023salient, li2023lightweight}.

\begin{figure}[t!]
    \centering
    \includegraphics[width=\linewidth]{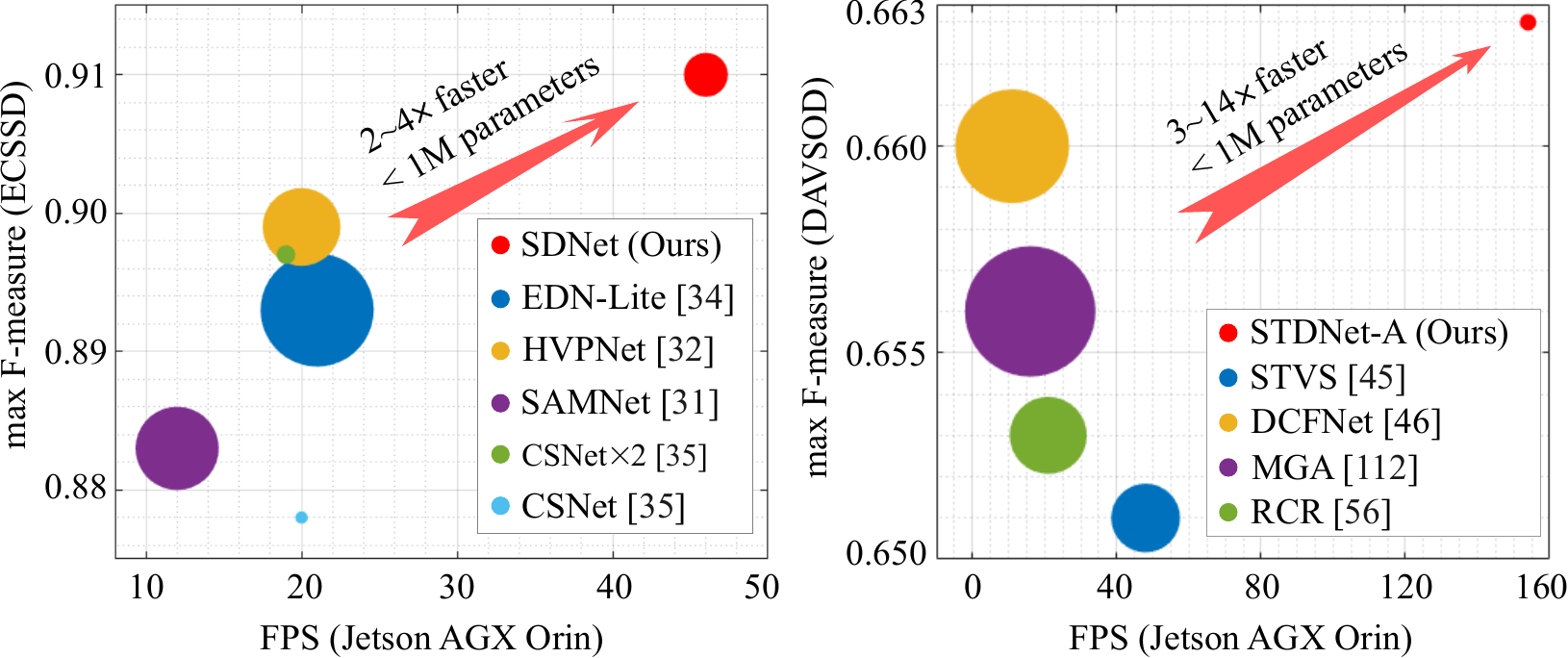}
    \caption{Comparing accuracy-runtime trade-offs for different methods on ISOD (left) and VSOD (right). All ISOD models are trained from scratch without pre-training. The circle size represents the number of parameters (illustrated separately for the two figures for better visualization). On both the ISOD and VSOD tasks, our proposed models achieve significantly better trade-offs with less than 1M parameters.
    }
    \label{fig:trade-off}
\end{figure}

Generally, SOD can be categorized into classical and deep learning approaches~\cite{borji2019sodsurvey}.
Classical methods segment salient regions based on handcrafted features and heuristics like uniqueness in spatial distributions, sparsity in low-rank representations, local and global contrasts in attributes like colors, orientations, and shapes~\cite{borji2019sodsurvey}. 
In contrast, later deep learning approaches use Deep Neural Networks (DNNs) like Convolutional Neural Networks (CNNs) ~\cite{kim2016cnncls1,wang2016cnncls2,kim2016cnncls1,wang2016cnncls2} and Vision Transformers (ViTs)~\cite{liu2021vst,zhuge2022icon} to obtain saliency maps. The problem of SOD has witnessed significant progress brought by DNNs due to their capability of learning multi-level image representations (from low-level details to high-level semantics) automatically from data, bypassing the need for feature engineering.

Mimicking the extraordinary ability of humans in detecting visually salient objects effortlessly 
and rapidly, SOD should be very efficient like the HVS in the first place, so that the majority of computational resources can be allocated for the down-stream tasks. 
Furthermore, with the ubiquitous use of mobile and embedded devices nowadays, such as IoTs and embedded systems that are under stringent resource limitations but in high demand of running latency among users, 
SOD need to be \emph{ultra-fast} so it does not become the bottleneck of the overall fast system. However, state-of-the-art SOD models tend to prioritize achieving progressively improved accuracy with intricate architectures or heavy backbones compromising on efficiency (see~\cref{tab:pretrain}).

Existing methods have tried to tackle this challenge with lightweight attention modules for multi-scale learning~\cite{liu2021samnet,liu2020hvpnet}, lightweight backbones~\cite{liu2022poolnet+,wu2022edn}, incorporating channel pruning to reduce feature redundancy~\cite{cheng2021csnet}, and adopting downsampling to obtain global views for salient object localization~\cite{wu2022edn}. Though lightweight in parameters, these methods suffer from suboptimal efficiency-accuracy trade-offs (see~\cref{fig:trade-off}),
due to the overly complex structures or constrained learning capacities. 

In this paper, we try to achieve fast runtime and high accuracy at the same time. 
We propose a method that leverages both, the heuristic insights from classical algorithms  and the representation power of CNNs. The resulting architecture is lightweight and efficient while achieving state-of-the-art performance on the SOD task. 
Typically, salient objects stand out due to their distinct visual cues, such as the contrasts in color, texture, or shape \cite{liu2010learning}. Thereby, the majority of classical SOD methods are based on \emph{contrast cues} in an image, \ie comparing pixels/regions of a neighboring area in feature spaces to acquire the uniqueness, distinctiveness, or rarity in a scene~\cite{borji2019sodsurvey,ma2003contrastacmmm,perazzi2012contrastcvpr,jiang2013ufo,cheng2013gc}. The insights behind these methods are from the mechanism of ``center-surround interaction'' for visual attention. The mechanism is well-studied in neuroscience where stimuli falling at positions in the surround modulate the response evoked by a stimulus appearing within the neuron’s classical receptive field (\ie the center)~\cite{sundberg2009center-surround,casagrande1991neural}, and first formulated in computer vision for saliency extraction by Itti \etal~\cite{itti1998itti} via sensing local spatial discontinuities. As discontinuities suggest the feature differences or contrasts among individual cells (\eg pixels, patches, or regions on the feature maps), saliency detection can be done by localizing those distinctive ones based on rich contrast patterns. 

Unlike classical methods using handcrafted pipelines to capture contrast cues, we leverage various Pixel Difference Convolutions (PDCs)~\cite{su2021pidinet} that help encode the center-surround relations and local discontinuities to achieve this goal. On the one hand, PDCs involve neighboring pixel comparison and extract high-order image cues like gradient statistics or feature disparities~\cite{pidinet-pami,su2021pidinet}. On the other hand, armed with a CNN architecture, the model is able to generate rich feature maps across various semantics and scales for multi-level (in semantic levels) and multi-scale feature contrast measuring. The calculation of contrast cues is embedded in the convolutional operators rather than through designing extra expensive modules. In addition, we develop a Difference Convolution Reparameterization (DCR) strategy to make PDCs free of parameters and computational overhead, by seamlessly integrating them into standard convolutional structures. All of these contributions allow us to build an effective and lightweight SOD model that ensures a high level of efficiency.

Besides the single Image SOD (ISOD), we further extend our method to videos (termed VSOD) via consideration of contrast cues in spatiotemporal feature spaces. To achieve that, we extend PDC to include SpatioTemporal Difference Convolutions (STDC) capable of capturing both spatial and motion cues. We design a LBP-TOP-style~\cite{lbp-top} 3D convolutional structure that decomposes the 3D spatiotemporal volume into two orthogonal time-space planes, on which our DCR is again applied without modification. 

Benchmarking on both consumer-grade GPUs and embedded systems, our models achieve considerably improved efficiency-accuracy trade-offs compared with existing state-of-the-art lightweight methods (\cref{fig:trade-off}). Our ISOD model runs at 252 and 46 FPS on a 2080 Ti GPU and an Nvidia AGX Orin embedded system, respectively, $4{\sim}6$ and $2{\sim}4$ times faster than existing lightweight competitors with similar accuracy. On VSOD, our model achieves 482 and 150 FPS on the above devices, running more than 3 times faster than competitors while achieving better prediction results.

While this work is related to our previous work that proposes PDCs~\cite{pidinet-pami}, it has several significant and independent contributions. While the previous work targets general vision tasks, the proposed one concentrates specifically on SOD with the following contribuions. (1) We incorporate the \emph{center-surround mechanism} into convolutional structures via PDCs to facilitate SOD. (2) We propose the novel DCR strategy to simplify PDC-based structures to basic standard convolutions, making PDCs free of computation and parameters during inference. (3) We develop new STDC operators for video saliency, compatible with DCR. (4) Our \emph{ultra-fast} and lightweight models demonstrate state-of-the-art performance with unprecented accuracy-runtime trade-offs on both ISOD and VSOD. 

The rest of the paper is organized as follows. \Cref{sec:related_work} introduces the related work, including lightweight SOD, approaches of enhancing model efficiency, and relation to PiDiNet~\cite{pidinet-pami,su2021pidinet} that also uses PDC operators. Then, our methods are presented in \cref{sec:methods} with detailed illustrations and discussions. In \cref{sec:experiments}, we investigate on several important questions regarding efficiency, memory, and data, and validate the effectiveness of our methods via comparative experiments. Finally, the paper is concluded in \cref{sec:conclusion}.

\section{Related Work}
\label{sec:related_work}

\noindent
\textcolor{black}{
\textbf{Lightweight SOD.}\quad
The research presented in this paper is part of the broader topic of SOD in computer vision, which encompasses various data modalities and involves the development of numerous lightweight architectures. These modalities include natural images~\cite{liu2021samnet,liu2020hvpnet,cheng2021csnet,wu2022edn}, videos~\cite{chen2021stvs,2021iccv-dcfnet}, remote sensing images~\cite{li2023lightweight}, and RGB-D data~\cite{wu2021mobilesal,zhang2021depth}. In this study, we specifically focus on the two widely used and significant modalities: natural images and videos.}

In the literature of \textbf{ISOD}, a great number of models have been proposed in recent years~\cite{wu2022edn,zhuge2022icon,liu2021vst,liu2019poolnet,zhao2019egnet}. With advances in deep learning, techniques like multi-scale feature fusion~\cite{zhang2017amulet,wang2018dgrl}, edge guided feature learning~\cite{zhao2019egnet,liu2019poolnet}, and feature attention~\cite{wang2019salientattention} in CNN architectures have improved the prediction accuracy of ISOD models. More recently, ViT-based models have brought new state-of-the-art results
by using global attention modules to capture any-distance
relationships among image regions~\cite{zhuge2022icon,liu2021vst}.

Specifically, efforts to achieve better accuracy-runtime trade-offs have resulted in many lightweight ISOD architectures~\cite{cheng2021csnet,liu2021samnet,liu2020hvpnet,wu2022edn}. SAMNet~\cite{liu2021samnet} designed a compact architecture and adopted a stereoscopic attention mechanism to automatically control the learning at different scales. To simulate the structure of the primate visual cortex in human brain, HVPNet~\cite{liu2020hvpnet} proposed a hierarchical visual perception (HVP) module by using the kernel sizes and dilation rates in descending order, based on which a lightweight ISOD model was designed. Meanwhile, EDN~\cite{wu2022edn} adopted an extremely downsampled block to learn a global view of the whole image and used MobileNetV2~\cite{sandler2018mobilenetv2} backbone to construct an efficient ISOD model. Recently, studies have shown that ISOD models require far fewer parameters than classification models and that ImageNet pre-training is not necessary for ISOD training~\cite{cheng2021csnet}. However, small models may not always result in low inference latency, since a more complicated model design might be harder to implement for runtime efficiency~\cite{ding2021repvgg}. For example, although there are only about 100K parameters in CSNet~\cite{cheng2021csnet}, we found it suffers from a suboptimal accuracy-efficiency trade-off due to its irregular layout in channel numbers (\eg, CSNet with 90K parameters runs at 61 FPS on the RTX 2080 Ti, while the ICON-R model~\cite{zhuge2022icon} with 33M parameters runs at 82 FPS). In our work, we develop a network that is more efficient in terms of compute, model size, and data usage, while still being equally or more accurate. 

In \textbf{VSOD}, temporal consistency and temporal saliency cues are equally critical as spatial cues. With deep learning, many VSOD methods are developed by leveraging temporal information in various ways, including via optical flow~\cite{2021iccv-dcfnet,yan2019rcr}, ConvLSTMs~\cite{2019cvpr-ssav,2018cvpr-fgrne,song2018pyramiddilated}, or 3D convolutions~\cite{chen2021stvs,le2017c3dbmvc}. More recently, long-term feature mining~\cite{chen2022limvsod}, multi-modal attention~\cite{lu2022depth}, and dynamic filters~\cite{2021iccv-dcfnet} have been proposed to further improve VSOD accuracy. However, relying on optical flow restricts the architecture to be end-to-end; the detection of optical flow also adds computational overhead, making the pipeline less suitable for real-time processing. ConvLSTM-based methods suffer from complex architectures with slow inference speed~\cite{2019cvpr-ssav}. While standard 3D convolutions run faster, their limited representational capacity may yield subpar prediction accuracy~\cite{chen2021stvs}.
Our method adopts the lightweight 3D convolutions, which however, are significantly augmented by our STDC operators that behave complementarily with the standard convolution to strengthen the overall model representational ability.
A similar work to ours is STVS~\cite{chen2021stvs}, which uses 3D convolutions for spatiotemporal feature encoding and aims at an architecture suitable for real-time performance. However, the standard convolution adopted in STVS limits its representational capacity, resulting in a suboptimal accuracy-runtime trade-off when compared with our proposed spatiotemporal network.

\begin{figure}[t!]
    \centering
    \includegraphics[width=\linewidth]{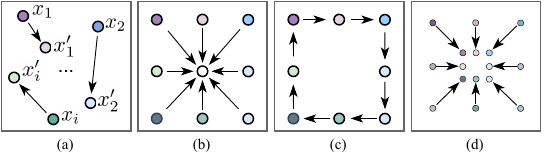}
    \caption{(a) Formulation for selecting pixel pairs in PDC; (b-d): specific selection strategies in CPDC, APDC, RPDC, respectively.
    }
    \label{fig:pdc}
\end{figure}

\vspace{0.3em}
\noindent \textcolor{black}{
\textbf{Model Efficiency.}\quad
Due to hardware constraints such as storage and computational limitations in real-word applications, deploying powerful deep learning architectures to fit such constraints is challenging. The recent works reduce computational costs through model efficiency methodologies, including compact architectures design \cite{chen2023run,fan2024lightweight,cai2023efficientvit,li2022efficientformer}, pruning \cite{su2020dynamic,he2023structured,ma2023llm,su2024boosting}, knowledge distillation (KD) \cite{gou2021knowledgedistill,hao2024one,li2023curriculum}, and quantization \cite{zhang2022dynamicthreshold,su2022svnet,li2024q,liu2023oscillation,xiao2023smoothquant}. Compact architectures design focuses on architectural-like optimizations, designing efficient convolution operations and model architectures to reduce computational overhead. Pruning involves reducing model size by eliminating redundant parameters, connections, or layers. KD aims to transfer the generalization and estimation capabilities from a stronger teacher model to a student model, \eg the lightweight models. Quantization maps weights and activations in models to a low-bit data format, effectively reducing memory requirements and accelerating model inference. In this paper, we focus on developing novel compact architectures for ISOD and VSOD. 
}

\vspace{0.3em}
\noindent \textcolor{black}{
\textbf{PiDiNet.}\quad
There might be certain shared insights between SDNet and PiDiNet~\cite{pidinet-pami,su2021pidinet} that both adopt PDC~\cite{su2021pidinet,zhang2022median} as the basic convolution operators. In this paper, we only treat PDC as a basic tool just like the standard convolution operator in various deep learning architectures. Like many CNNs using standard convolution, we established new values for PDC in the SOD literature. First of all, how to prevent SOD from becoming a latency bottleneck that hinders the ``SOD + downstream application'' system from achieving real-time performance is a key research topic, especially considering that the downstream application may itself be time-intensive. Different to the researches in the SOD literature focusing on accuracy while ignoring efficiency, we demonstrated that the PDC-based architectures can achieve much better efficiency-accuracy trade-offs compared with prior approaches. Second, this paper is the pioneering work to make PDC parameter- and computation-free, though straightforward, it is important to investigate whether reparameterization of ``PDC + standard convolution'' or the traditional ``standard convolution + standard convolution'' makes effect on the task of SOD. As shown in our ablation studies, the existing routine of using ``standard convolution + standard convolution'' reparameterization strategy fails to give extra accuracy gain. In contrast, we demonstrated that the PDC-based reparameterization indeed improve the model performance due to their complementarity to standard convolution. Last, the extension from 2D PDC to 3D STDC allows explicit feature contrast capturing in spatiotemporal spaces that goes beyond the standard 3D convolution, which therefore benefits VSOD. 
}

\section{Methods}
\label{sec:methods}

\begin{figure}[t!]
    \centering
    \includegraphics[width=\linewidth]{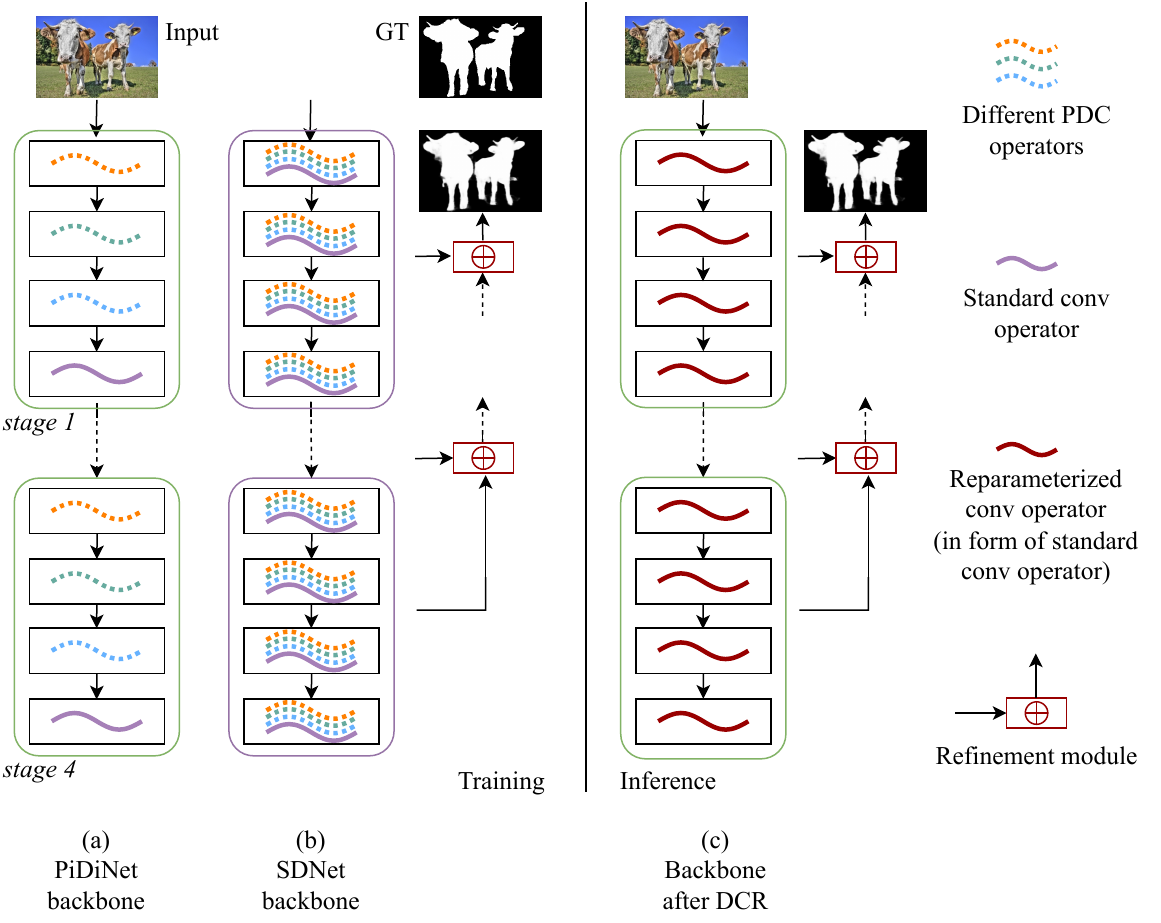}
    \caption{Architecture overview. (a) PiDiNet backbone~\cite{su2021pidinet}; (b) SDNet backbone during training; (c) SDNet backbone during inference. Best viewed in color.
    }
    \label{fig:framework}
\end{figure}

\begin{figure*}[t!]
    \centering
    \includegraphics[width=\linewidth]{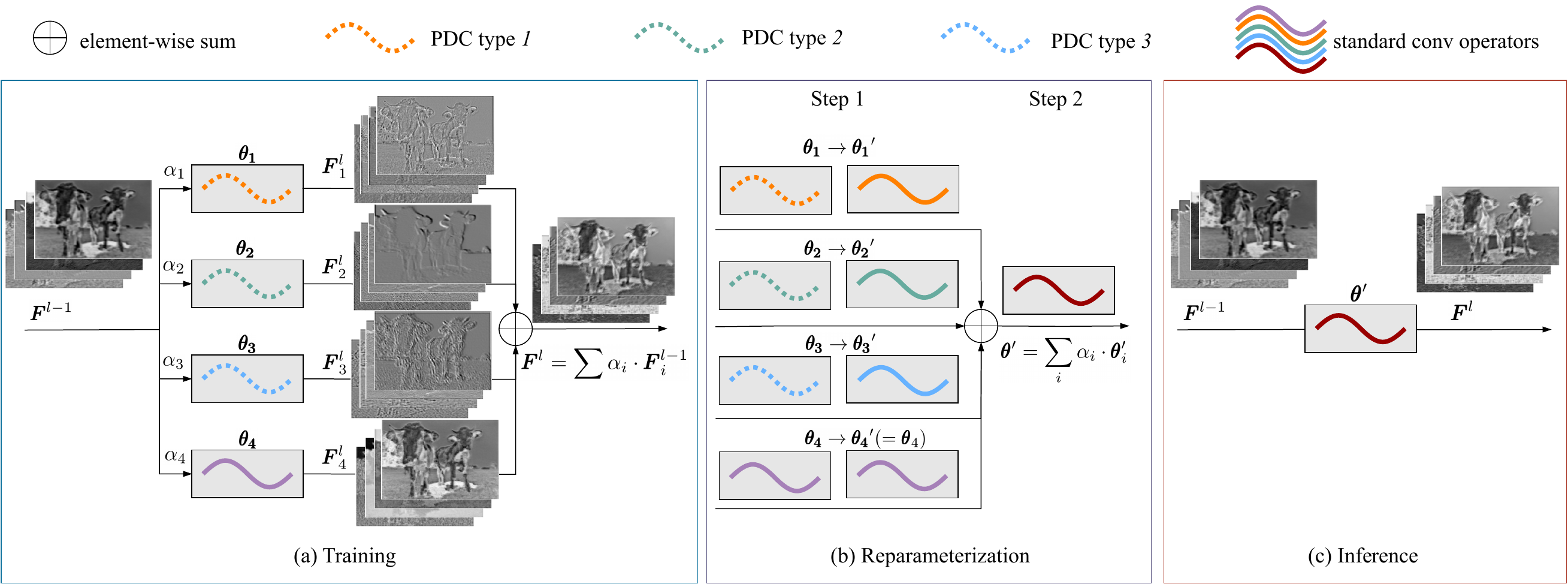}
    \caption{Our proposed DCR pipeline. In this example, we employ three different PDC operators and a standard convolutional operator. However, any number of PDC operators can be considered without affecting the final efficiency. Best viewed in color.
    }
    \label{fig:rep}
\end{figure*}

\subsection{Pixel Difference Convolution (PDC) Revisited}
\label{sec:method-pdc}

In standard convolutions, an output value is computed from the inner product between kernel weights and pixel intensities in a local region of the feature map. Differently, PDC initially selects pixel pairs within a local region, and then computes the inner product between kernel weights and the pixel differences between pairs. Supposing that the current local region $R$ consists of $\pmb{x}^R=\{x_1^R, x_2^R, ..., x_n^R\}$ with $n$ pixels, the standard convolution and PDC are formulated as:

\begin{align}
    y^R &= f(\pmb{x}^R, \pmb{\theta}) = \sum_{i=1}^{n}w_i\cdot x_i^R, \;\;\;\;\; \text{(Standard convolution)} \\
    y^R &= f(\Delta\pmb{x}^R, \pmb{\theta}) = \sum_{i=1}^{m}w_i\cdot (x_i^R - x_i'^R), \;\;\;\;\;\, \text{(PDC)} \label{eq: pdc}
\end{align}
where $y^R$ is the output value at the center of region $R$, $\pmb{\theta} = \{w_1, w_2, ..., w_i, ...\}$ are the kernel weights, $(x_i^R, x_i'^R)$ is a pair of pixels selected from $\pmb{x}^R$ (\ie $x_i^R, x_i'^R\in\pmb{x}^R$), and $m$ is the number of pixel pairs.

Originated from LBP~\cite{ojala2002lbp},  PDC is an ``abstract operator'' that can have different forms via different strategies of selecting pixel pairs. For instance, the Central PDC (CPDC), Angular PDC (APDC), and Radial PDC (RPDC) are created by selecting pixel pairs along the central, angular, and radial directions respectively~\cite{pidinet-pami}, as shown in~\cref{fig:pdc}. The rich ways to build various forms of PDC operators enable the model to capture rich contrast patterns in feature maps. As the feature maps are generated across different scales and semantic levels due to the use of a CNN architecture, the PDC-based saliency extraction remarkably goes beyond those classical SOD approaches using basic features like color, intensity, orientations, and so on. 

The use of PDCs is in line with the well-studied frequency domain interpretation in the SOD literature~\cite{itti1998itti,achanta2009frequency} since they act like high-pass filters~\cite{pidinet-pami}. Dating back to~\cite{reinagel1999natural}, Reinagel and Zador found the spatial frequency content at the fixated locations to be significantly higher than, on average, at random locations. Since eye fixation shows high correlation with object saliency~\cite{borji2019sodsurvey}, Itti\etal~\cite{itti1998itti} was able to reproduce the above findings via computing the differences of Gaussian-smoothed images in a frequency pyramid. This suggests that saliency detection is coupled with the extraction of high-frequency signals. In addition, retaining high frequencies is facilitative to generate saliency maps with well-defined object boundaries~\cite{achanta2009frequency}. 

Based on the above insights, we need various PDC operators at each layer of our target CNN architecture to extract rich multi-level and multi-scale contrast cues, and retain high frequencies. Meanwhile, we also need the standard convolution operators to preserve the fundamental low-frequency components~\cite{pidinet-pami} (\eg for highlighting large salient objects~\cite{achanta2009frequency}). This leads to our backbone structure where multiple types of convolution operators are jointly applied at each layer.

\subsection{SDNet - A Solution for ISOD}

To this end, we propose Spatial Difference Network (SDNet) for the ISOD task. SDNet is based on PiDiNet~\cite{pidinet-pami}, with the difference that PiDiNet employs only one type of operator at each backbone layer which is suitable for the task of edge detection~\cite{pidinet-pami}, but limits its representational capacity for SOD. Instead, ours supports an arbitrary number of convolutional operators with different types at single layers (\cref{fig:framework}), including different PDCs and the standard convolution. Thereby, both high-order contrast patterns and zeroth-order intensity cues are explicitly encoded for an enhanced feature representation. Based on that, coefficients are learned for the operators at each layer to automatically aggregate their contributions. To avoid a proportional increase in number of parameters and computational cost during inference, we introduce our Difference Convolution Reparameterization (DCR) strategy that fuses the convolution operators into a single operator. Therefore, the backbone is converted to a plain architecture identical to that of PiDiNet but has a higher representation power due to the coupled use of different convolution types. 

\vspace{0.3em}
\noindent \textbf{Difference Convolution Reparameterization (DCR).}\quad
During training (\cref{fig:rep} (a)), for each layer in the SDNet backbone, we employ different PDC types as well as the standard convolution via individual convolutional branches and thereby the high-order contrast cues and zeroth-order intensities are both considered. At the end of this layer, we take the weighted sum of the outputs from the branches to fuse these different features. The contributions of features are learned by a set of coefficients $\{\alpha_i\}$ assigned to the branches.

\begin{table*}[t!]
\caption{Backbone configuration of SDNet and SDNet-A. ``Conv'' means standard convolution. $3\times 3$ or $5\times 5$ represents the kernel size. Since that a $3\times 3$ RPDC is converted to a $5\times 5$ standard convolution~\cite{pidinet-pami}, we only adopt it in specific layers to save computation. Please refer to MobileViTv2~\cite{mehta2022mobilevitv2} for the structure of the attention block.}
\centering
\setlength{\tabcolsep}{0.008\linewidth}
\resizebox*{\linewidth}{!}{
\begin{tabular}{c|c|c|c|c|c}
\toprule
Layer & Output & \multicolumn{2}{c}{SDNet backbone} & \multicolumn{2}{c}{SDNet-A backbone} \\
\cmidrule(r){3-6}
& & Training & Inference & Training & Inference \\
\midrule
1 & $H\times W$ & $3\times 3$ (Conv + CPDC + APDC), 60 & $3\times 3$ Conv, 60 & $3\times 3$ (Conv + CPDC + APDC), 60 & $3\times 3$ Conv, 60\\
2 & $H/2\times W/2$ & $3\times 3$ (Conv + CPDC + APDC), 60 & $3\times 3$ Conv, 60 & $3\times 3$ (Conv + CPDC + APDC), 60 & $3\times 3$ Conv, 60\\
3 & $H/2\times W/2$ & $3\times 3$ (Conv + CPDC + APDC), 60 & $3\times 3$ Conv, 60 & $3\times 3$ (Conv + CPDC + APDC), 60 & $3\times 3$ Conv, 60\\
4 & $H/2\times W/2$ & $3\times 3$ (Conv + CPDC + APDC + RPDC), 60 & $5\times 5$ Conv, 60 & $3\times 3$ (Conv + CPDC + APDC + RPDC), 60 & $5\times 5$ Conv, 60\\
\midrule
5 & $H/4\times W/4$ & $3\times 3$ (Conv + CPDC + APDC), 120 & $3\times 3$ Conv, 120 & $3\times 3$ (Conv + CPDC + APDC), 120 & $3\times 3$ Conv, 120\\
6 & $H/4\times W/4$ & $3\times 3$ (Conv + CPDC + APDC), 120 & $3\times 3$ Conv, 120 & $3\times 3$ (Conv + CPDC + APDC), 120 & $3\times 3$ Conv, 120\\
7 & $H/4\times W/4$ & $3\times 3$ (Conv + CPDC + APDC), 120 & $3\times 3$ Conv, 120 & Attention Block, 120 & Attention Block, 120\\
8 & $H/4\times W/4$ & $3\times 3$ (Conv + CPDC + APDC + RPDC), 120 & $5\times 5$ Conv, 120 & Attention Block, 120 & Attention Block, 120\\
\midrule
9 & $H/8\times W/8$ & $3\times 3$ (Conv + CPDC + APDC), 240 & $3\times 3$ Conv, 240 & $3\times 3$ (Conv + CPDC + APDC), 240 & $3\times 3$ Conv, 240\\
10 & $H/8\times W/8$ & $3\times 3$ (Conv + CPDC + APDC), 240 & $3\times 3$ Conv, 240 & $3\times 3$ (Conv + CPDC + APDC), 240 & $3\times 3$ Conv, 240\\
11 & $H/8\times W/8$ & $3\times 3$ (Conv + CPDC + APDC), 240 & $3\times 3$ Conv, 240 & Attention Block, 240 & Attention Block, 240\\
12 & $H/8\times W/8$ & $3\times 3$ (Conv + CPDC + APDC + RPDC), 240 & $5\times 5$ Conv, 240 & Attention Block, 240 & Attention Block, 240\\
\midrule
13 & $H/16\times W/16$ & $3\times 3$ (Conv + CPDC + APDC), 240 & $3\times 3$ Conv, 240 & $3\times 3$ (Conv + CPDC + APDC), 240 & $3\times 3$ Conv, 240\\
14 & $H/16\times W/16$ & $3\times 3$ (Conv + CPDC + APDC), 240 & $3\times 3$ Conv, 240 & $3\times 3$ (Conv + CPDC + APDC), 240 & $3\times 3$ Conv, 240\\
15 & $H/16\times W/16$ & $3\times 3$ (Conv + CPDC + APDC), 240 & $3\times 3$ Conv, 240 & Attention Block, 240 & Attention Block, 240\\
16 & $H/16\times W/16$ & $3\times 3$ (Conv + CPDC + APDC + RPDC), 240 & $5\times 5$ Conv, 240 & Attention Block, 240 & Attention Block, 240\\
\bottomrule
\end{tabular}
}
\label{tab:backbone}
\end{table*}

After training (\cref{fig:rep} (b)), for each PDC operator $i$ with weight kernel $\pmb{\theta}_i$, following~\cite{su2021pidinet}, we reparameterize $\pmb{\theta}_i$ to $\pmb{\theta}_i'$ to convert PDC to standard convolution (step 1):

\textcolor{black}{
\begin{align}
    &f(\Delta\pmb{x}, \pmb{\theta}_i) = \sum_{j}w_{i,j}\cdot(x_j - x_j')\nonumber\\
    &= \sum_jx_j\cdot(w_{i,j} - \sum_{k\in\mathcal{Q}_j}w_{i,k}) = \sum_jx_j\cdot w_{i,j}' = f(\pmb{x}, \pmb{\theta}_i'),
\end{align}
}
where $\mathcal{Q}_j$ gathers the coefficients of ``$-x_j$'' in the equation. 

Then, we reparameterize all the parameters $\{\alpha_i, \pmb{\theta}_i'\}$ in the current layer to $\pmb{\theta}'$ to convert the multi-branch inference to a single-branch version (step 2):

\begin{align}
    &y = \sum_i \alpha_i\cdot f(\Delta\pmb{x}, \pmb{\theta}_i) = \sum_i \alpha_i\cdot f(\pmb{x}, \pmb{\theta}_i'))\nonumber\\
    &=\sum_i \alpha_i\cdot \sum_j x_j\cdot w_{i,j}' = \sum_j x_j\cdot \sum_i \alpha_i\cdot w_{i,j}'\nonumber\\
    &=f(\pmb{x}, \sum_i \alpha_i\cdot \pmb{\theta}_i') = f(\pmb{x}, \pmb{\theta}')
\end{align}

By doing so, the layers are transformed into a plain PiDiNet architecture (\cref{fig:rep} (c)).

\begin{figure}[t!]
    \centering
    \includegraphics[width=\linewidth]{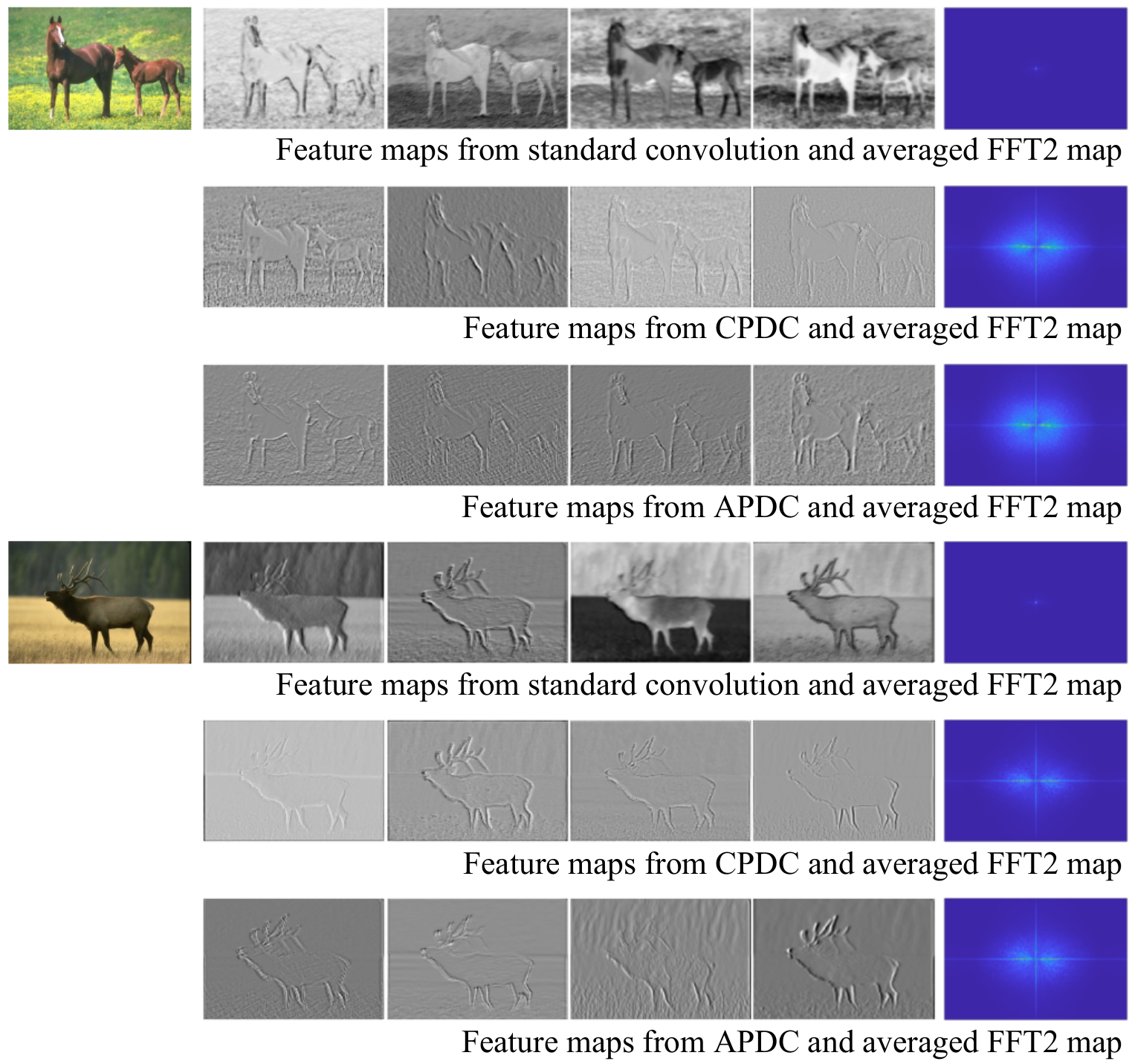}
    \caption{We visualize the intermediate feature maps in layer 4 of SDNet. The averaged FFT2 map for each row is obtained by averaging the FFT2 maps of all the feature maps in the output, followed by normalization to [0, 1].
    }
    \label{fig:fft2}
\end{figure}

\vspace{0.3em}
\noindent \textbf{Network Structures of SDNet.}\quad
Leveraging DCR, it is possible to configure each backbone layer with arbitrary number of convolutional operators with a constant computational cost at inference time. We utilize the three basic PDCs from PiDiNet and the standard convolution as illustrated in \cref{tab:backbone}. \textcolor{black}{It should be noticed that since RPDC is converted to $5\times 5$ convolutions, it costs more computation and memory than CPDC and APDC. Therefore, we only adopt it at the last layer of each backbone stage to strike a balance between representation and efficiency.}

With the backbone generating multi-stage features, the side architectures (other parts of the network rather than the backbone) aim to produce a full-resolution saliency map. Following~\cite{pidinet-pami}, the feature maps generated by the four stages of the backbone are processed by the Compact Dilation Convolution Module (CDCM) and Compact Spatial Attention Module (CSAM)~\cite{pidinet-pami} to reduce the channels, denoted as $\{\pmb{O}_k| k= 1, 2, 3, 4\}$.   
We then adopt an efficient top-down feature refinement pipeline to gradually refine the output feature maps from the four stages of the backbone. Let $\pmb{F}_k\in \mathbb{R}^{C_k\times H_k\times W_k}$ be the refined feature maps of stage $k$ where $C_k$, $H_k$, and $W_k$ represent its number of channels, height, and width respectively. We first upsample $\pmb{F}_k$ to match the size of $\pmb{O}_{k-1}$ via linear interpolation, denoted as $\pmb{F}_k'$, then concatenate $\pmb{F}_k'$ and $\pmb{O}_{k-1}$. After that, we use a $3\times 3$ convolution to reduce the number of channels of the concatenated features to $C_{k-1}$ and get $\pmb{F}_{k-1}$. Note that for the last stage, $\pmb{F}_4 = \pmb{O}_4$. The final saliency map is obtained by a linear transformation based on $\pmb{F}_1$ followed by linear interpolation to recover the original input size.

In addition, since vision transformers (ViT) with a global attention mechanism trained on large-scale data have achieved promising results in various computer vision tasks including the dense prediction ones, we further investigate whether the global attention mechanism can benefit ISOD with our constraints: small model size, real-time inference, and limited training data.
We integrate the lightweight attention module in MobileViTv2~\cite{mehta2022mobilevitv2} into the backbone by replacing the last two convolutional layers in each of the last three stages of SDNet backbone with two attention blocks, leading to the SDNet-A (``A'' means attention) backbone. We choose the attention module from MobileViTv2 for two reasons: first, it is highly compact and efficient; second, it is scalable to large image resolutions since its computational cost is linear to the number of tokens. SDNet-A backbone is also illustrated in~\cref{tab:backbone}.

\vspace{0.3em}
\noindent \textbf{Discussion: DCR as a reparameterization strategy.}\quad
Network reparameterization~\cite{ding2021repvgg,guo2020expandnets} facilitates the learning of simpler structures which may show similarities with knowledge distillation~\cite{gou2021knowledgedistill}: the final inference-phrase structure is much simpler than the training-phrase version or the teacher network, which is hard to be that powerful when trained alone. Integration of PDCs promotes the learning of contrast cues for a simple standard convolutional structure. 

From the view of frequency domain, prior reparameterization methods fuse convolutional operators in the spatial space (either by fusing operators with different kernel shapes or fusing consecutive layers), while DCR enables a single operator to explicitly fuse low- and high-frequency features. 
As analyzed in~\cite{pidinet-pami}, PDCs perform in a way that are more likely to highlight high-frequency features,  while the standard convolution is prone to maintaining the original frequency components from the input, which are dominated by low frequencies. Some intermediate feature maps and corresponding FFT2 results are shown in~\cref{fig:fft2}, where we can see high-frequency signals are significantly emphasized by PDC. With DCR, although the inference-time structure only contains standard convolution operators, the impacts of PDCs for highlighting high frequencies are still preserved. In \cref{sec:ablation}, we have also shown that simply reparameterizing multiple standard convolutional operators fails to give performance gain in the case of SOD.

\subsection{STDNet - An Extension to VSOD}
Once the ISOD problem is tackled by SDNet in the image domain, an important concern raises: can the spirits of PDC and DCR apply to the video domain to tackle the VSOD problem? In this part, we extend our methodology to videos and target a lightweight model for real-time VSOD with a novel spatiotemporal convolutional operator. The new operator is an extension of PDC from the spatial space to the spatiotemporal space, which considers both motion and appearance information to better learn high-order contrast patterns in videos. The operator is also made computation- and parameter-free since it is well compatible with DCR. 

\begin{figure}[t!]
    \centering
    \includegraphics[width=\linewidth]{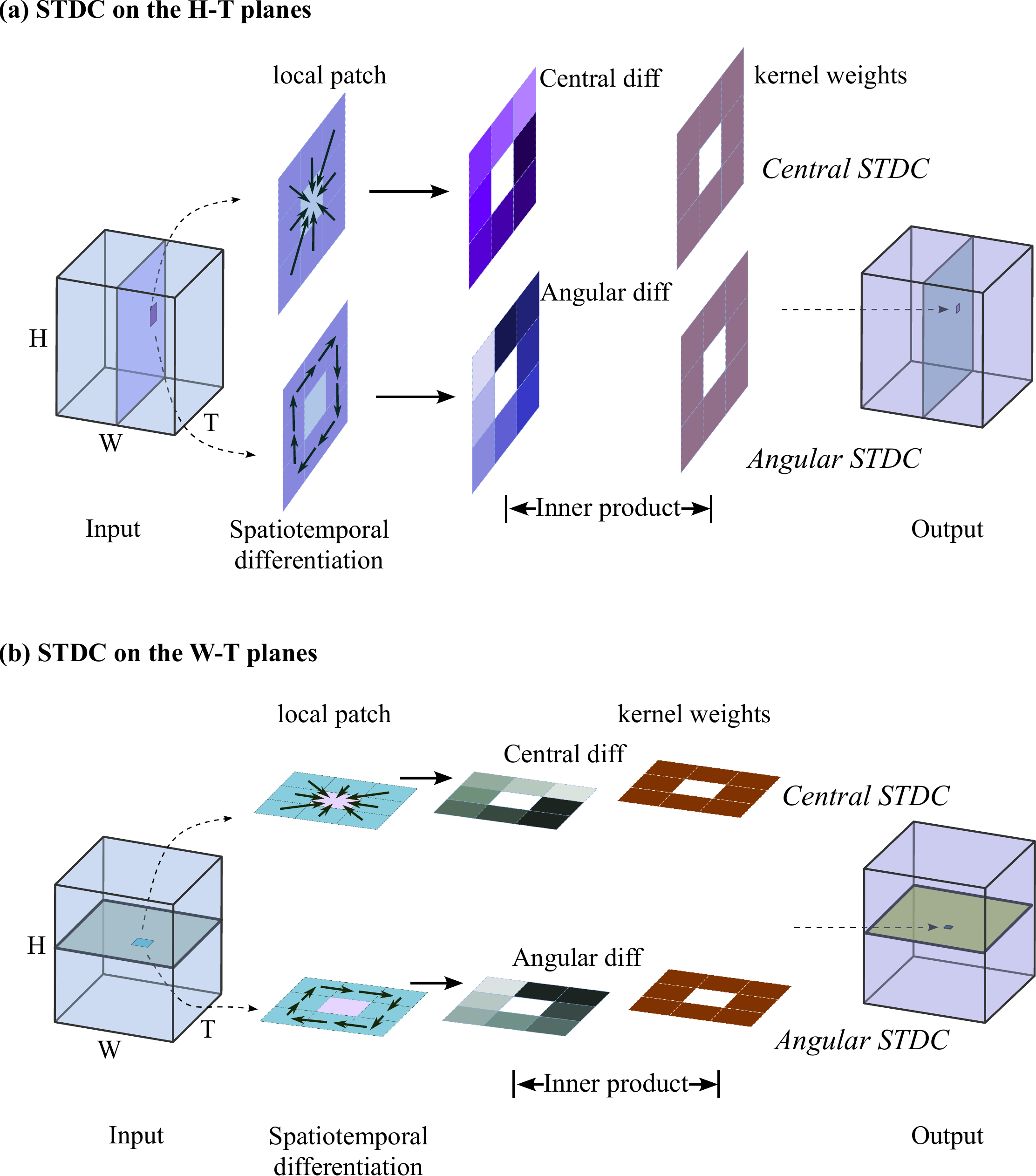}
    \caption{Illustration of the proposed STDC in H-T and W-T planes.
    }
    \label{fig:stdc}
\end{figure}

\vspace{0.3em}
\noindent \textbf{SpatioTemporal Difference Convolution (STDC).}\quad
Standard 3D convolution uses pixel intensities in a local 3D region to probe spatiotemporal patterns. Nonetheless, the zeroth-order intensities might not explicitly represent local temporal contrast in the frame sequence, a factor that could be pivotal for the model's assessment of temporal consistency in consistently detecting salient objects across video frames.
Similar to PDCs' capacity of encoding spatial contrast, STDC is designed to measure feature contrast in spatiotemporal space, by expanding the feature comparison to both time and spatial dimensions. Particularly, the pixel pairs are selected across different spatial and temporal locations and thereby the patterns captured by STDC reflect both motion and appearance dynamics. To ease computation, instead of designing a 3D operator that selects pixel pairs among the whole 3D input volume, we adopt a simpler approximation where the 3D volume is sliced into W-T and H-T planes separately, and deploy 2D PDC-like convolutions respectively, as illustrated in \cref{fig:stdc}.  

Interpretably, the W-H plane is the original image space. 
While the W-T plane gives a visual impression of one row changing in time and H-T describes the motion of one column in temporal space (please see \cref{fig:tdc_features}). Both planes are thus considered. The orthogonal slicing benefits the implementation of STDC by converting its formulation to that of PDC in~\cref{eq: pdc} without efforts, where the local region $R$ is replaced with the W-T or H-T slice in the local 3D region.

\begin{figure}[t!]
    \centering
    \includegraphics[width=0.8\linewidth]{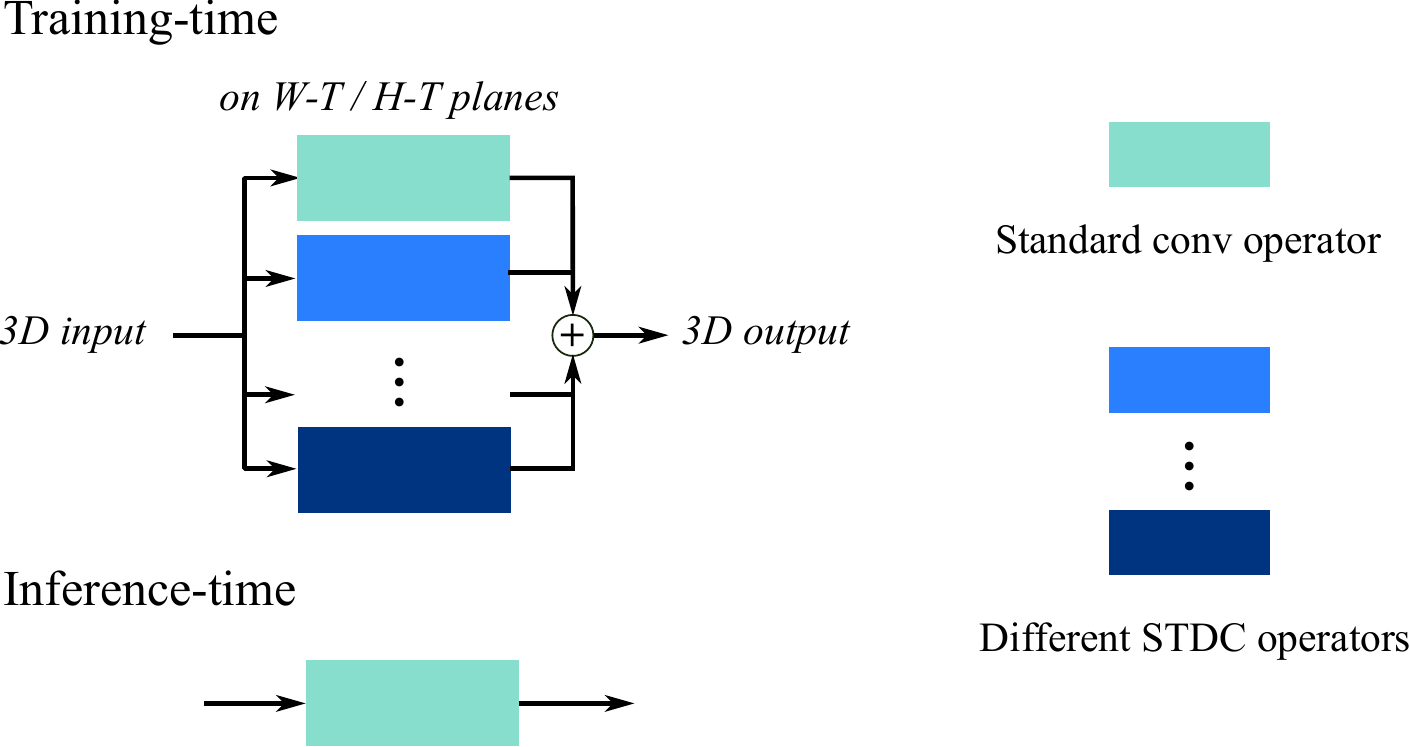}
    \caption{Illustration of a STDC layer with DCR that can be executed on either W-T or H-T planes with arbitrary number of operators.
    }
    \label{fig:slice_stdc}
\end{figure}

\begin{figure*}[t!]
    \centering
    \includegraphics[width=\linewidth]{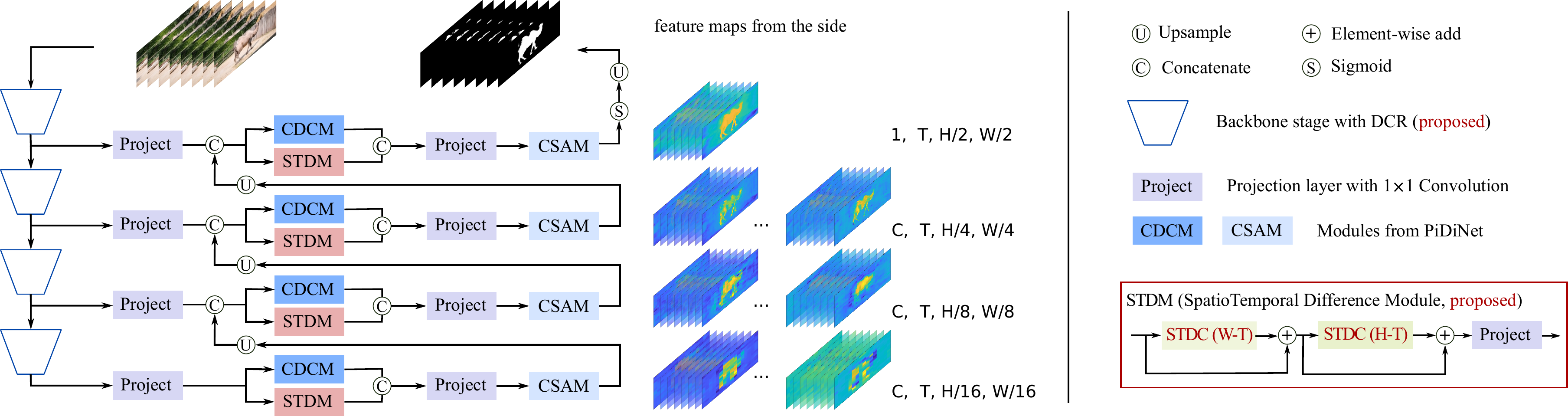}
    \caption{The proposed STDNet architecture. STDC (W-T) and STDC (H-T) are implemented following \cref{fig:slice_stdc}.
    }
    \label{fig:stdnet}
\end{figure*}

Analogous to PDC, STDC serves as an ``abstract operator'' that can be instantiated with different forms by using certain pixel pair selection strategies on W-T or H-T planes. We visualize Central STDC (CSTDC) and Angular STDC (ASTDC) in~\cref{fig:stdc} (a) \& (b) for H-T and W-T planes. Once we get different types of STDC operators, we could similarly build a multi-branch structure during training, noting that the standard convolution operator is also adopted for preserving low frequencies, following SDNet (\cref{fig:slice_stdc}). 
With our DCR strategy after training, we can easily transform the multi-branch structure into a single-branch version consisting of only standard convolution (reparameterized); this renders the involved STDC operators as computation- and parameter-free.

\vspace{0.3em}
\noindent \textbf{Network Structures of STDNet.}\quad
The overall architecture is shown in~\cref{fig:stdnet}. The basic idea is to reuse our SDNet backbone to capture spatial information and leave the side structures to generate the final saliency map by further incorporating spatiotemporal saliency cues. \textcolor{black}{The reason of keeping the backbone unchanged is to maximize compatibility between the two SOD tasks, such that a single pretrained backbone can be used for both SDNet and STDNet.} For the side architecture, we develop our SpatioTemporal Difference Module (STDM), where two consecutive STDC-based layers are conducted on W-T and H-T planes respectively to cover the whole 3D input. In our implementation, we adopt ASTDC, CSTDC, and the standard convolutional operators to build our STDC layers; an ablation study is given in~\ref{sec:ablation}. The side structures serve two purposes: (1) spatial and temporal feature refinement, and (2) multi-stage feature aggregation. At each stage, we use CDCM~\cite{su2021pidinet} and our STDM for spatial and spatiotemporal feature refinement respectively. The features from these two modules are then aggregated via concatenation, and further processed by CSAM~\cite{su2021pidinet} for background suppression. To encourage multi-stage feature aggregation, we use a top-down approach for refining our feature maps. Specifically, the output of CSAM at each stage is concatenated with the backbone-extracted features from the previous stage prior to being fed into the spatial and temporal refinement modules again.

\vspace{0.3em}
\noindent \textbf{Discussion: STDC \vs LBP-TOP.}\quad
Like PDC being a learnable LBP descriptor for images in the spatial domain~\cite{pidinet-pami}, the proposed STDC behaves as a learnable LBP-TOP descriptor~\cite{lbp-top} for videos when additionally considering the temporal dimension. Similar design spirit was present in LBP-TOP where three orthogonal planes (XY, XT, and YT\footnote{X, Y, and T corresponds to W, H, and T in our method.}) were separated from the 3D local volume to calculate their corresponding co-occurrences of neighboring pixels (\ie LBP patterns). The time-space feature representation was then generated by concatenating the statistics of the local patterns from these three planes. As a non-learnable descriptor, LBP-TOP has fixed number of possible patterns like that in LBP. For instance, setting the number of neighboring points to $p$ for each plane gives $3\cdot 2^p$ bins when calculating the global feature histogram of a given $X\times Y\times T$ dynamic texture. The reason behind its non-learnability is that LBP-TOP only takes the signs of the pixel differences and uses binomial factors as internal kernel weights to calculate features. While the proposed STDC preserves the values of pixel differences and allows the kernel weights to be learned from data. Taking the W-T plane (\ie XT plane) as an example, assuming the central pixel as $g_c$ and its neighboring pixels as $\{g_0, g_1, ..., g_p\}$, the formulations of LBP-TOP and central STDC are presented as:

\begin{align}
    f &= \sum_{i=0}^p2^i\cdot \text{Sign}(g_i - g_c), \;\;\; \text{(for LBP-TOP)}\\
    f &= \sum_{i=0}^pw_i\cdot (g_i - g_c), \;\;\; \text{(for central STDC)}
\end{align}
where $f$ is the calculated feature and $\{w_i\}$ are learnable parameters. 

\section{Experiments}
\label{sec:experiments}

In our experiments, we focus on investigating the following questions:
\begin{itemize}
    \item As our SDNet and STDNet are designed to be suitable for resource-limited devices, how lightweight are our models and how do they perform on real hardware targets?
    \item Regarding the prediction performance on ISOD and VSOD, which trade-off do SDNet and STDNet achieve between efficiency and accuracy?
    \item Cheng \etal~\cite{cheng2021csnet} claimed that ImageNet pretraining is not necessary for lightweight ISOD models since they have adequate capacity to capture necessary salient semantics. Does this apply to SDNet and STDNet for ISOD and VSOD respectively? How can we achieve a better balance across efficiency, accuracy, and amount of labeled data?
    \item How does each of the design choices, such as our DCR strategy, the fusion of PDCs, and the STDC type contribute to model performance?
\end{itemize}

For the first three, we focus on factors like inference speed, memory consumption, accuracy, and data labeling that are challenging the real-world applications where resources are limited 
(\eg IoTs and embedded systems). For the last question, we aim to give a clear ablation study for our models.

\begin{table*}[t!]
\caption{Comparison with prior models on ISOD without pre-trained backbones. 
The best results are marked \textbf{in bold}. EffFormer, MBViT, and MBViTv2 are abbreviations of EfficientFormer, MobileViT, and MobileViTv2 respectively. \textcolor{black}{$S$, $F$, and $M$ indicates the $S_\lambda(\uparrow)$, $F_\beta^m(\uparrow)$, and MAE$(\downarrow)$ metrics respectively.}
Re-implemented models are denoted with $^\dagger$. Please refer to the text for more details.}
\centering
\setlength{\tabcolsep}{0.006\linewidth}
\resizebox*{\linewidth}{!}{
\begin{tabular}{lc>{\color{black}}cccc>{\color{black}}ccc|>{\color{black}}ccc|>{\color{black}}ccc|>{\color{black}}ccc|>{\color{black}}ccc|>{\color{black}}ccc}
\toprule
\multirow{2}{*}{Models} & \multirow{2}{*}{\begin{tabular}{@{}c@{}}\#Param\\(M)\end{tabular}} & \multirow{2}{*}{\begin{tabular}{@{}c@{}}FLOPs\\(G)\end{tabular}} & \multirow{2}{*}{\begin{tabular}{@{}c@{}}FPS\\(2080 Ti)\end{tabular}} & \multirow{2}{*}{\begin{tabular}{@{}c@{}}FPS\\(Orin)\end{tabular}} & \multirow{2}{*}{\begin{tabular}{@{}c@{}}Input\\size\end{tabular}} & \multicolumn{3}{c}{ECSSD} & \multicolumn{3}{c}{PASCAL-S} & \multicolumn{3}{c}{DUT-O} & \multicolumn{3}{c}{HKU-IS} & \multicolumn{3}{c}{SOD} & \multicolumn{3}{c}{DUTS-TE} \\
& & & &  & & $S$ & $F$ & $M$ & $S$ & $F$ & $M$ & $S$ & $F$ & $M$ & $S$ & $F$ & $M$ & $S$ & $F$ & $M$ & $S$ & $F$ & $M$ \\
\midrule
\multicolumn{17}{l}{Models for salient object detection} \\
\midrule
DSS$^\dagger$~\cite{HouPami19dss} & 62.24 & 196.52 & 48 & 9 & 320$^2$ & .881 & .884 & .065 & .818 & .798 & .094 & .801 & .724 & .075 & .874 & .870 & .056 & .738 & .754 & .138 & .824 & .774 & .067 \\
EDN-R$^\dagger$~\cite{wu2022edn} & 42.85 & 28.30 & 52 & 14 & 320$^2$ & .888 & .899 & .056 & .795 & .776 & .103 & \textbf{.815} & .743 & .068 & .886 & .887 & .047 & .730 & .753 & .143 & .826 & .773 & .067 \\
ICON-R$^\dagger$~\cite{zhuge2022icon} & 33.09 & 34.51 & 82 & 18 & 320$^2$ & .819 & .804 & .089 & .737 & .695 & .123 & .697 & .586 & .095 & .786 & .758 & .084 & .654 & .646 & .166 & .727 & .635 & .092 \\
\textcolor{black}{ICON-S}$^\dagger$~\cite{zhuge2022icon} & 92.40 & 105.33 & 38 & 8 & 384$^2$ & .818 & .813 & .093 & .747 & .709 & .126 & .735 & .624 & .094 & .810 & .786 & .080 & .674 & .681 & .173 & .741 & .650 & .096 \\
\midrule
\multicolumn{17}{l}{\colorbox{gray!30}{Lightweight models} for other dense prediction tasks} \\
\midrule
ESPNetV2$^\dagger$~\cite{mehta2019espnetv2} & 0.34 & 0.60 & 118 & 35 & 320$^2$ & .881 & .886 & .071 & .803 & .780 & .108 & .799 & .723 & .079 & .869 & .865 & .063 & .735 & .757 & .146 & .812 & .756 & .076 \\
PiDiNet$^\dagger$~\cite{su2021pidinet} & 0.72 & 4.02 & 275 & 46 & 320$^2$ & .893 & .897 & .055 & .813 & .799 & .093 & .803 & .728 & .068 & .883 & .881 & .050 & .762 & .779 & .123 & .830 & .783 & \textbf{.061} \\
BiSeNetV2$^\dagger$~\cite{yu2021bisenetv2} & 3.34 & 9.58 & 191 & 51 & 320$^2$ & .891 & .900 & .057 & .807 & .790 & .097 & .809 & .736 & .067 & .885 & .884 & .048 & .738 & .758 & .134 & .830 & .781 & .062 \\
ENet$^\dagger$~\cite{paszke2016enet} & 0.35 & 1.52 & 106 & 27 & 320$^2$ & .896 & .901 & .060 & .824 & .807 & .095 & \textbf{.815} & .747 & .075 & .887 & .886 & .053 & .755 & .770 & .134 & .832 & .787 & .068 \\
DABNet$^\dagger$~\cite{li2019dabnet} & 0.75 & 4.02 & 154 & 48 & 320$^2$ & .897 & .902 & .056 & \textbf{.826} & .808 & .088 & .813 & .739 & .068 & \textbf{.892} & .891 & .047 & .752 & .782 & .129 & \textbf{.838} & .790 & .060 \\
\midrule
\multicolumn{17}{l}{\colorbox{gray!30}{Lightweight models} (ViT models) on classification} \\
\midrule
\textcolor{black}{EffFormer}$^\dagger$~\cite{li2022efficientformer} & 11.57 & 2.70 & 166 & 47 & 224$^2$ & .875 & .881 & .063 & .792 & .774 & .105 & .788 & .713 & .080 & .866 & .863 & .056 & .732 & .753 & .142 & .800 & .742 & .075 \\
\textcolor{black}{EdgeNeXt}$^\dagger$~\cite{maaz2022edgenext} & 1.19 & 0.85 & 170 & 45 & 320$^2$ & .875 & .877 & .060 & .794 & .772 & .104 & .805 & .730 & .070 & .872 & .868 & .052 & .729 & .744 & .143 & .812 & .756 & .068 \\
\textcolor{black}{MBViT}$^\dagger$~\cite{DBLP:MehtaR22/mobilevit} & 0.96 & 1.09 & 119 & 31 & 320$^2$ & .883 & .887 & .061 & .797 & .774 & .107 & .802 & .732 & .078 & .881 & .877 & .051 & .732 & .739 & .142 & .815 & .758 & .072 \\
\textcolor{black}{MBViTv2}$^\dagger$~\cite{mehta2022mobilevitv2} & 1.17 & 1.78 & 131 & 38 & 320$^2$ & .865 & .865 & .069 & .785 & .760 & .109 & .793 & .717 & .079 & .874 & .870 & .053 & .719 & .728 & .148 & .805 & .746 & .073 \\
\midrule
\multicolumn{17}{l}{\colorbox{gray!30}{Lightweight models} for salient object detection}\\
\midrule
CSNet~\cite{cheng2021csnet} & 0.09 & 0.44 & 61 & 20 & 224$^2$ & .877& .878 & .076 & .802 & .783 & .112 & .795 & .724 & .087 & .870 & .867 & .066 & .744 & .766 & .149 & .808 & .757 & .082 \\
CSNet$\times 2$~\cite{cheng2021csnet} & 0.14 & 0.72 & 60 & 19 & 224$^2$ & .893 & .897 & .066 & .813 & .797 & .104 & .805 & .737 & .080 & .882 & .881 & .060 & .756 & .781 & .137 & .822 & .779 & .074 \\
SAMNet~\cite{liu2021samnet} & 1.33 & 0.95 & 39 & 12 & 320$^2$ & .878 & .883 & .072 & .804 & .780 & .109 & .811 & .736 & .075 & .872 & .863 & .064 & .733 & .748 & .149 & .817 & .752 & .076 \\
HVPNet~\cite{liu2020hvpnet} & 1.24 & 2.01 & 47 & 20 & 320$^2$ & .886 & .899 & .066 & .806 & .796 & .104 & .813 & \textbf{.754} & .074 & .881 & .884 & .058 & .731 & .771 & .142 & .823 & .778 & .071 \\
EDN-Lite~\cite{wu2022edn} & 1.80 & 1.42 & 65 & 21 & 320$^2$ & .894 & .893 & .061 & .802 & .785 & .098 & .810 & .745 & \textbf{.068} & .885 & .882 & .049 & .735 & .758 & .143 & .827 & .776 & .065 \\
\midrule
\midrule
\textcolor{black}{SDNet-A} & 0.82 & 3.56 & 169 & 36 & 320$^2$ & .893 & .900 & .055 & .802 & .784 & .101 & .796 & .734 & .077 & .884 & .889 & .048 & .765 & .751 & .136 & .814 & .770 & .072 \\
SDNet & 0.71 & 3.12 & 252 & 46 & 320$^2$ & \textbf{.899} & \textbf{.910} & \textbf{.052} & .823 & \textbf{.816} & \textbf{.087} & .803 & .739 & .070 & .888 & \textbf{.897} & \textbf{.046} & \textbf{.778} & \textbf{.786} & \textbf{.118} & .828 & \textbf{.798} & .063 \\
\bottomrule
\end{tabular}
}
\label{tab:scratch}
\end{table*}

\begin{table*}[t!]
\caption{Comparison with prior models on ISOD with ImageNet pre-trained backbones.  
To get the metrics for prior ISOD methods, we download the predicted saliency maps released by the authors and test them via the same evaluation code for a fair comparison. If the original method utilized the actual image sizes, then the speed is evaluated with the resolution of $320\times 320$. \textcolor{black}{$S$, $F$, and $M$ indicates the $S_\lambda(\uparrow)$, $F_\beta^m(\uparrow)$, and MAE$(\downarrow)$ metrics respectively.}}
\centering
\setlength{\tabcolsep}{0.006\linewidth}
\resizebox*{\linewidth}{!}{
\begin{tabular}{lc>{\color{black}}cccc>{\color{black}}ccc|>{\color{black}}ccc|>{\color{black}}ccc|>{\color{black}}ccc|>{\color{black}}ccc|>{\color{black}}ccc}
\toprule
\multirow{2}{*}{Models} & \multirow{2}{*}{\begin{tabular}{@{}c@{}}\#Param\\(M)\end{tabular}} & \multirow{2}{*}{\begin{tabular}{@{}c@{}}FLOPs\\(G)\end{tabular}} & \multirow{2}{*}{\begin{tabular}{@{}c@{}}FPS\\(2080 Ti)\end{tabular}} & \multirow{2}{*}{\begin{tabular}{@{}c@{}}FPS\\(Orin)\end{tabular}} & \multirow{2}{*}{\begin{tabular}{@{}c@{}}Input\\size\end{tabular}} & \multicolumn{3}{c}{ECSSD} & \multicolumn{3}{c}{PASCAL-S} & \multicolumn{3}{c}{DUT-O} & \multicolumn{3}{c}{HKU-IS} & \multicolumn{3}{c}{SOD} & \multicolumn{3}{c}{DUTS-TE} \\
& & & &  & & $S$ & $F$ & $M$ & $S$ & $F$ & $M$ & $S$ & $F$ & $M$ & $S$ & $F$ & $M$ & $S$ & $F$ & $M$ & $S$ & $F$ & $M$ \\
\midrule
UCF~\cite{zhang2017ucf} & 29.47 & 146.42 & - & - & - & .883 & .890 & .069 & .806 & .791 & .114 & .760 & .698 & .120 & .875 & .874 & .062 & .763 & .773 & .148 & .782 & .742 & .111 \\
Amulet~\cite{zhang2017amulet} & 33.15 & 40.22 & - & - & - & .894 & .905 & .059 & .819 & .810 & .098 & .781 & .715 & .098 & .886 & .887 & .051 & .755 & .765 & .144 & .804 & .750 & .084\\
SRM~\cite{wang2017srm} & 53.14 & 36.82 & - & - & 353$^2$ & .895 & .905 & .054 & .834 & .822 & .083 & .798 & .725 & .069 & .887 & .893 & .046 & .746 & .791 & .126 & .836 & .797 & .058 \\
DGRL~\cite{wang2018dgrl} & 161.74 & 191.28 & - & - & 384$^2$ & .903 & .914 & .041 & .836 & .832 & .072 & .806 & .739 & .062 & .895 & .900 & .036 & .774 & .800 & .103 & .842 & .805 & .050 \\
PoolNet~\cite{liu2019poolnet} & 68.26 & 153.60 & 52 & 8 & actual & .926 & .937 & .035 & .865 & .858 & .065 & .831 & .763 & .054 & .919 & .923 & .030 & .792 & .830 & .104 & .887 & .865 & .036 \\
EGNet~\cite{zhao2019egnet} & 111.69 & 488.28 & 19 & 4 & actual & .925 & .936 & .037 & .852 & .846 & .074 & .841 & .778 & .053 & .918 & .924 & .031 & .807 & .844 & .097 & .887 & .865 & .039 \\ 
ICON-R~\cite{zhuge2022icon} & 33.09 & 41.77 & 80 & 17 & 352$^2$ & .929 & .943 & .032 & .861 & .865 & .064 & .844 & .799 & .057 & .920 & .930 & .029 & .824 & .850 & .084 & .889 & .876 & .037 \\
EDN-R~\cite{wu2022edn} & 42.85 & 40.72 & 53 & 13 & 384$^2$ & .927 & .941 & .033 & .864 & .865 & .062 & .850 & .799 & .050 & .924 & .932 & .027 & - & - & - & .892 & .878 & .035 \\
\textcolor{black}{VST}~\cite{liu2021vst} & 44.56 & 23.16 & 45 & 9 & 224$^2$ & .932 & .944 & .034 & .873 & .850 & .067 & .850 & .800 & .058 & .928 & .937 & .030 & .854 & .866 & .065 & .896 & .877 & .037 \\
\textcolor{black}{ICON-S}~\cite{zhuge2022icon} & 92.40 & 105.33 & 38 & 8 & 384$^2$ & .941 & .954 & .023 & .884 & .888 & .048 & .869 & .830 & .043 & .935 & .947 & .022 & .825 & .859 & .083 & .917 & .910 & .024 \\
\midrule
\midrule
SAMNet~\cite{liu2021samnet} & 1.33 & 1.06 & 39 & 12 & 336$^2$ & .907 & .915 & .051 & .826 & .811 & .092 & .830 & .773 & .065 & .898 & .901 & .045 & .762 & .792 & .124 & .849 & .811 & .057 \\
HVPNet~\cite{liu2020hvpnet} & 1.24 & 2.21 & 48 & 19 & 336$^2$ & .904 & .912 & .053 & .830 & .815 & .090 & .831 & .773 & .064 & .899 & .902 & .045 & .765 & .793 & .123 & .849 & .814 & .057 \\
EDN-lite~\cite{wu2022edn} & 1.80 & 2.04 & 64 & 20 & 384$^2$ & .911 & .923 & .043 & .842 & .835 & .073 & .824 & .757 & .058 & .907 & .912 & .034 & - & - & - & .862 & .835 & .045 \\
\midrule
SDNet & 0.71 & 4.50 & 248 & 35 & 384$^2$ & .898 & .908 & .052 & .824 & .816 & .088 & .798 & .733 & .073 & .888 & .896 & .047 & .778 & .780 & .117 & .830 & .801 & .062 \\
\textcolor{black}{SDNet-A} & 0.82 & 5.12 & 165 & 26 & 384$^2$ & .908 & .922 & .045 & .820 & .815 & .089 & .818 & .773 & .067 & .903 & .915 & .038 & .788 & .798 & .113 & .839 & .816 & .057 \\
\bottomrule
\end{tabular}
}
\label{tab:pretrain}
\end{table*}

\begin{figure*}[t!]
    \centering
    \includegraphics[width=\linewidth]{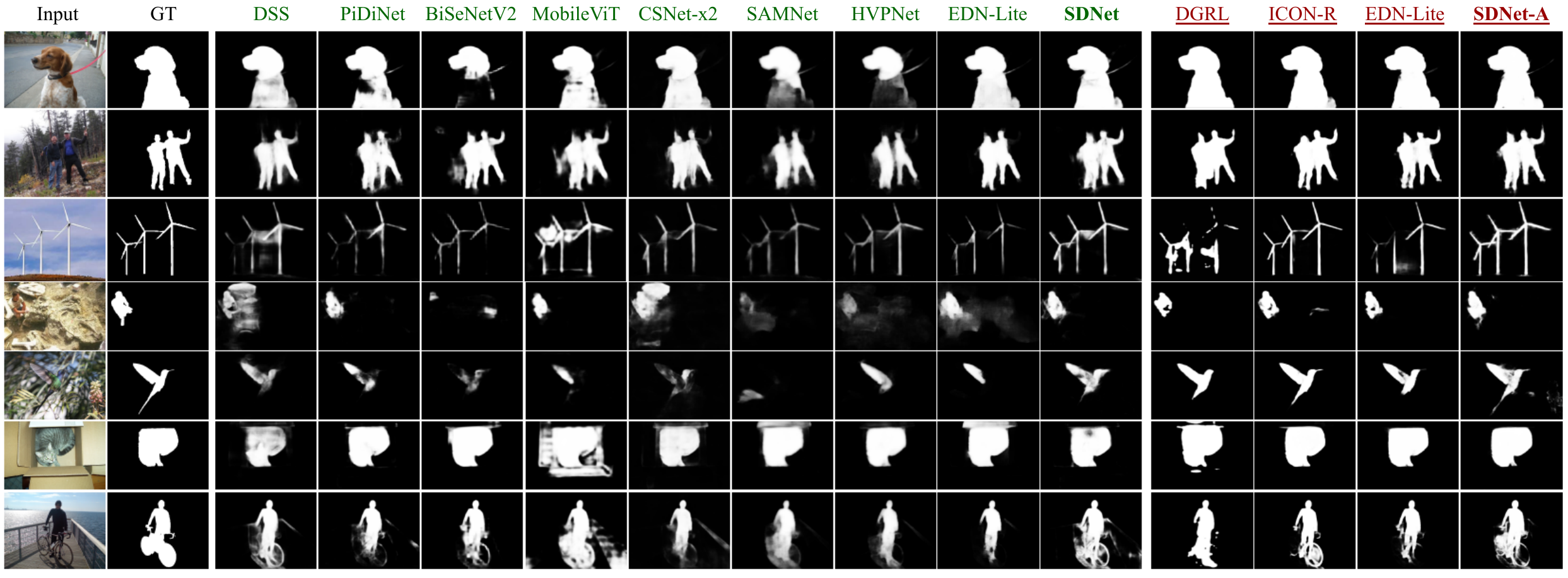}
    \caption{Qualitative comparison on ISOD. The first two columns are the input images and corresponding ground truth images respectively. Other columns contain saliency maps from different models. We mark model A as ``\textcolor{mygreen}{A}'' or ``\textcolor{myred}{\underline{A}}'' if it was trained w/o or w/ pre-trained backbones.
    }
    \label{fig:quality}
\end{figure*}

\subsection{Experimental settings}

\noindent \textbf{Datasets.}\quad
We use the DUTS-TR~\cite{wang2017duts}, DUTS-TE~\cite{wang2017duts}, ECSSD~\cite{yan2013ecssd}, PASCAL-S~\cite{li2014pascals}, DUT-O~\cite{yang2013dutomron}, SOD~\cite{movahedi2010sod} and HKU-IS~\cite{li2015hkuis} datasets to evaluate on ISOD. Following previous works~\cite{cheng2021csnet,wu2022edn,liu2021samnet,zhuge2022icon}, we train our models on DUTS-TR and evaluate performance on the other datasets. To evaluate on VSOD, we use the DAVSOD~\cite{2019cvpr-ssav}, VOS~\cite{li2018vos}, and DAVIS~\cite{perazzi2016davis} datasets. Both DAVSOD and VOS have a validation set. For DAVIS, we randomly choose 8 videos from its training set as validation set. Our models are then evaluated on the corresponding test sets after training.

\vspace{0.3em}
\noindent \textbf{Implementation.}\quad
For ISOD, our models are trained for 180 epochs with Adam optimizer~\cite{kingma2014adam} and an initial learning rate of 0.001 that is decayed by 0.1 at epochs 90 and 150; the batch size is set to 24. We scale the input images to $320\times 320$ and interpolate the output images back to the original sizes. We use two RTX 3090 GPUs to train the models with Pytorch~\cite{paszke2019pytorch}. A weighted binary cross-entropy loss~\cite{liu2019richer} is adopted during training.

For VSOD, we follow prior works~\cite{2021iccv-dcfnet,chen2021stvs} by first removing the temporal modules (STDM) and training the rest of the model architectures on ISOD data (DUTS-TR) with 60 epochs (stage 1), then fine-tuning with a combination of VSOD (using the training sets from the three VSOD datasets) and ISOD data (DUTS-TR) with another 60 epochs (stage 2). If the input is from the ISOD data in stage 2, then a ``boring'' video clip from a single image is created via duplication following~\cite{chen2021stvs}. For both stages of training, the learning rate is initialized with 0.001, and decayed at epoch 30 and 50 respectively, with a decaying rate of 0.1. It should be noted that the initial learning rate of the backbone is further reduced by 0.01 in the second stage. However, when the backbone is already pretrained with ImageNet~\cite{deng2009imagenet}, we skip stage 1 and directly conduct stage 2 with the combined data. For each input clip, 8 frames are used with a resolution of $256\times 256$. Since 8 is much less than 256 in the spatial dimensions, we adopt replicate padding on the temporal dimension when conducting convolution in STDM following~\cite{chen2021stvs}. The batch size is set to 8.
Like the denotation of SDNet-A, we denote the variant of STDNet with ViT blocks inserted into the backbone as STDNet-A (SDNet-A and STDNet-A share the same backbone architecture).
All models are implemented with Pytorch~\cite{paszke2019pytorch} and trained on two RTX 3090 GPUs. We adopt
the same loss with~\cite{2021iccv-dcfnet}, which includes the binary cross
entropy loss $L_{bce}$, IOU Loss $L_{IoU}$~\cite{yu2016iouloss} and SSIM Loss
$L_{ssim}$~\cite{wang2004ssimloss}. The final loss $L$ can
be expressed as $L = L_{bce} + L_{IoU} + L_{ssim}$. \textcolor{black}{During inference, the saliency maps are generated every 8 frames sequentially without overlapping. When the number of total frames is not divisible by 8, we left pad the last remaining frames to get a 8-frame clip input.}

\vspace{0.3em}
\noindent \textbf{Evaluation metrics.}\quad
For efficiency, we compare our models with prior ones in model size, \textcolor{black}{amount of floating point operations (FLOPs),} and real inference speed computed with batch size 1 (to simulate streamed data in real-time interactive applications) on the following devices: a popular consumer-grade GPU (Nvidia RTX 2080 Ti) and an embedded system (Nvidia Jetson AGX Orin). The speed is recorded in the form of frames per second (FPS). We additionally evaluate our VSOD models on the even more resource-limited embedded device Nvidia Xavier NX to further test the speed limit of our models. For accuracy, we report \textcolor{black}{the structure measure value ($S_\lambda$)~\cite{fan2017structure},} maximum F-measure score ($F_\beta^{m}$, with $\beta^2=0.3$) and mean absolute error (MAE) of ISOD models as is common in the literature~\cite{cheng2021csnet,wu2022edn,liu2021vst,liu2021samnet}. Note that there are two different ways to calculate F-measure score $F_\beta$: (1) calculate the averaged precision and recall over the complete dataset, based on which the overall $F_\beta$ is obtained~\cite{cheng2021csnet,wu2022edn}; (2) calculate $F_\beta$ values for individual images and take the average over the dataset~\cite{liu2021vst, zhuge2022icon}. While (1) often results in higher $F_\beta$ values than (2), we adopt (2) for all comparisons in this paper following the latest survey~\cite{wang2021sodsurvey}. For VSOD, we consistently take $S_\lambda$, $F_\beta^m$ ($\beta^2=0.3$), and MAE following the previous works~\cite{2021iccv-dcfnet,2019cvpr-ssav,chen2021stvs}. The metrics on VSOD are calculated with the toolbox used in~\cite{2019cvpr-ssav}.

\begin{table*}[t!]
\caption{Comparison with prior models on VSOD. $\star$ denotes traditional methods. FPS numbers marked with $\dagger$ are cited from~\cite{chen2022limvsod} (calculated on 2080 Ti). FPS numbers marked with $\ddagger$ are cited from the original papers, which were calculated on different GPUs other than 2080 Ti. \textcolor{black}{FLOPs are averaged for a single video frame.}}
\centering
\setlength{\tabcolsep}{0.008\linewidth}
\resizebox*{\linewidth}{!}{
\begin{tabular}{lc>{\color{black}}ccccccccc|ccc|ccc}
\toprule
Model & \multirow{2}{*}{\begin{tabular}{@{}c@{}}\#Params\\(M)\end{tabular}} & \multirow{2}{*}{\begin{tabular}{@{}c@{}}FLOPs\\(G)\end{tabular}} & \multirow{2}{*}{\begin{tabular}{@{}c@{}}FPS\\(2080 Ti)\end{tabular}} & \multirow{2}{*}{\begin{tabular}{@{}c@{}}FPS\\(Orin)\end{tabular}} & \multirow{2}{*}{\begin{tabular}{@{}c@{}}FPS\\(NX)\end{tabular}} & \multirow{2}{*}{\begin{tabular}{@{}c@{}}Input\\size\end{tabular}} & \multirow{2}{*}{\begin{tabular}{@{}c@{}}Backbone\\Pre-Train\end{tabular}} & \multicolumn{3}{c}{DAVSOD (2019)} & \multicolumn{3}{c}{VOS (2018)} & \multicolumn{3}{c}{DAVIS (2016)} \\
& & & & & & & & $S_\lambda\uparrow$ & $F_\beta^{m}\uparrow$ & MAE$\downarrow$ & $S_\lambda\uparrow$ & $F_\beta^{m}\uparrow$ & MAE$\downarrow$ & $S_\lambda\uparrow$ & $F_\beta^{m}\uparrow$ & MAE$\downarrow$ \\
\midrule
MSTM$^\star$~\cite{2016cvpr-mstm} & - & - & - & - & - & - & \xmark & .532 & .344 & .211 & .657 & .567 & .144 & .583 & .429 & .165 \\
SCOM$^\star$~\cite{2018tip-scom} & - & - & - & - & - & - & \xmark & .599 & .464 & .220 & .712 & .690 & .162 & .832 & .783 & .048 \\
\midrule
FGRNE~\cite{2018cvpr-fgrne} & - & - & - & - & - & 256$\times$512 & \checkmark & .693 & .573 & .098 & .715 & .669 & .097 & .838 & .783 & .043 \\
RCR~\cite{yan2019rcr} & 53.79 & 223.17 & 42 & 21 & 1.9 & 448$^2$ & \checkmark & .741 & .653 & .087 & .873 & .833 & .051 & .886 & .848 & .027 \\
SSAV~\cite{2019cvpr-ssav} & - & - & 20$^\dagger$ & - & - & 473$^2$ & \checkmark & .724 & .603 & .092 & .819 & .742 & .073 & .893 & .861 & .028 \\
MGA~\cite{li2019mga} & 91.52 & 246.50 & 33 & 16 & 1.6 & 512$^2$ & \checkmark & .751 & .656 & .081 & .792 & .735 & .075 & .912 & .892 & .022 \\
EG-GCN~\cite{2021tip-eg-gcn} & - & - & 0.5$^\ddagger$ & - & - & - & - & - & - & - & - & - & - & .880 & .844 & .031 \\
DCFNet~\cite{2021iccv-dcfnet} & 71.66 & 188.38 & 30 & 11 & 1.5 & 448$^2$ & \checkmark & .741 & .660 & .074 & .846 & .791 & .060 & .914 & .900 & .016 \\
CFCN-MA~\cite{zheng2022cfcn-ma} & - & - & 27$^\ddagger$ & - & - & 224$^2$ & \checkmark & .712 & .568 & .085 & - & - & - & .888 & .867 & .020 \\
LIMVSOD~\cite{chen2022limvsod} & - & -& 2.4$^\dagger$ & - & - & 352$^2$ & \checkmark & .792 & .725 & .064 & .844 & .822 & .060 & .922 & .911 & .016 \\
\midrule
\midrule
PCSA~\cite{gu2020pcsa} & 2.63 & - & 116 & - & - & 256$\times$448 & \checkmark & .741 & .655 & .086 & .827 & .747 & .065 & .902 & .880 & .022 \\
STVS~\cite{chen2021stvs} & 48.23 & 38.27 & 107 & 48 & 6.6 & 256$^2$ & \checkmark & .746 & .651 & .086 & .850 & .791 & .058 & .892 & .865 & .023 \\
\midrule
STDNet & 0.87 & 4.09 & 542 & 162 & 28 & 256$^2$ & \xmark & .703 & .579 & .094 & .835 & .789 & .071 & .869 & .837 & .029 \\
STDNet-A & 0.99 & 4.37 & 482 & 150 & 27 & 256$^2$ & \checkmark & .755 & .663 & .087 & .852 & .799 & .065 & .884 & .858 & .024 \\
\bottomrule
\end{tabular}
}
\label{tab:vsod}
\end{table*}

\subsection{Comparison with state-of-the-art methods}
\label{sec:experiments-state-of-the-art}

\noindent \textbf{Comparison on ISOD.}\quad
In line with the work in~\cite{cheng2021csnet}, we expect our SDNet to meet all the following needs towards our goals on ISOD: fast, memory-friendly, trainable from scratch using limited labeled data (without ImageNet pretraining). For a comprehensive validation, we consider prior models including (1) the latest state-of-the-art large-scale models~\cite{HouPami19dss,wu2022edn,zhuge2022icon}, (2) well-known lightweight CNN and ViT models designed for other dense prediction/classification tasks~\cite{mehta2019espnetv2,su2021pidinet,yu2021bisenetv2,paszke2016enet,li2019dabnet}, and (3) recent lightweight ISOD models~\cite{cheng2021csnet,liu2021samnet,liu2020hvpnet,wu2022edn}. All of these models are trained from scratch like SDNet and SDNet-A. 

For DSS~\cite{HouPami19dss} and ICON~\cite{zhuge2022icon}, we use the code provided by the authors to construct the models and then integrate them into our code for training with the same scheme. For CSNet~\cite{cheng2021csnet}, which was also trained from scratch on the same dataset like ours, we directly evaluate it with the predicted saliency maps released by the authors. For SAMNet~\cite{liu2021samnet}, HVPNet~\cite{liu2020hvpnet}, and EDN~\cite{wu2022edn}, we run the training scripts provided by the authors and slightly tune the learning rate schedules, since the original learning rates were designed for pre-trained backbones. For models on other tasks, we transfer them for the ISOD task following~\cite{cheng2021csnet} and adopt the same training configuration as ours. More specifically, when transferring the lightweight ViT models (designed for classification) to the ISOD task, we use our top-down feature refinement module to gradually increase the resolution of feature maps, resulting in a final saliency map. For segmentation models, the number of channels of the output layer is changed from the number of classes to 1. \textcolor{black}{In these settings, we are able to control the resolutions of the inputs. Hence, we adopt a single resolution for fair comparison (except ICON-S~\cite{zhuge2022icon} due to its specific ViT version that only accepts fixed resolutions) and CSNet series~\cite{cheng2021csnet} that were already trained from scratch with a specific resolution.}  Quantitative results are compared in~\cref{tab:scratch}. A qualitative comparison can be found in~\cref{fig:quality}.

All the lightweight models have only around 1M parameters. We find that if trained from scratch, those lightweight models for other dense prediction tasks behave as strong baselines for the ISOD task, since they share similar design principles with the ISOD counterparts: multi-scale feature fusion and refinement. However, \cref{tab:scratch} clearly indicates that the proposed SDNet achieves a much better trade-off between efficiency and accuracy. It demonstrates the superiority of the specifically designed SDNet architecture on ISOD task, \ie its additional capacity to extract high-order contrast cues for salient objects. 
For example, comparing with the latest lightweight ISOD models, SDNet achieves a $F_\beta^m$ score of 0.910 on ECSSD \vs 0.899 by the next best model (HVPNet), while running $5\times$ and $2\times$ faster during inference on the RTX 2080 Ti and the AGX Orin respectively. \Cref{fig:quality} also shows that our models detect object boundaries more clearly than others due to an explicit highlighting of high frequencies.

\begin{figure}[t!]
    \centering
    \includegraphics[width=\linewidth]{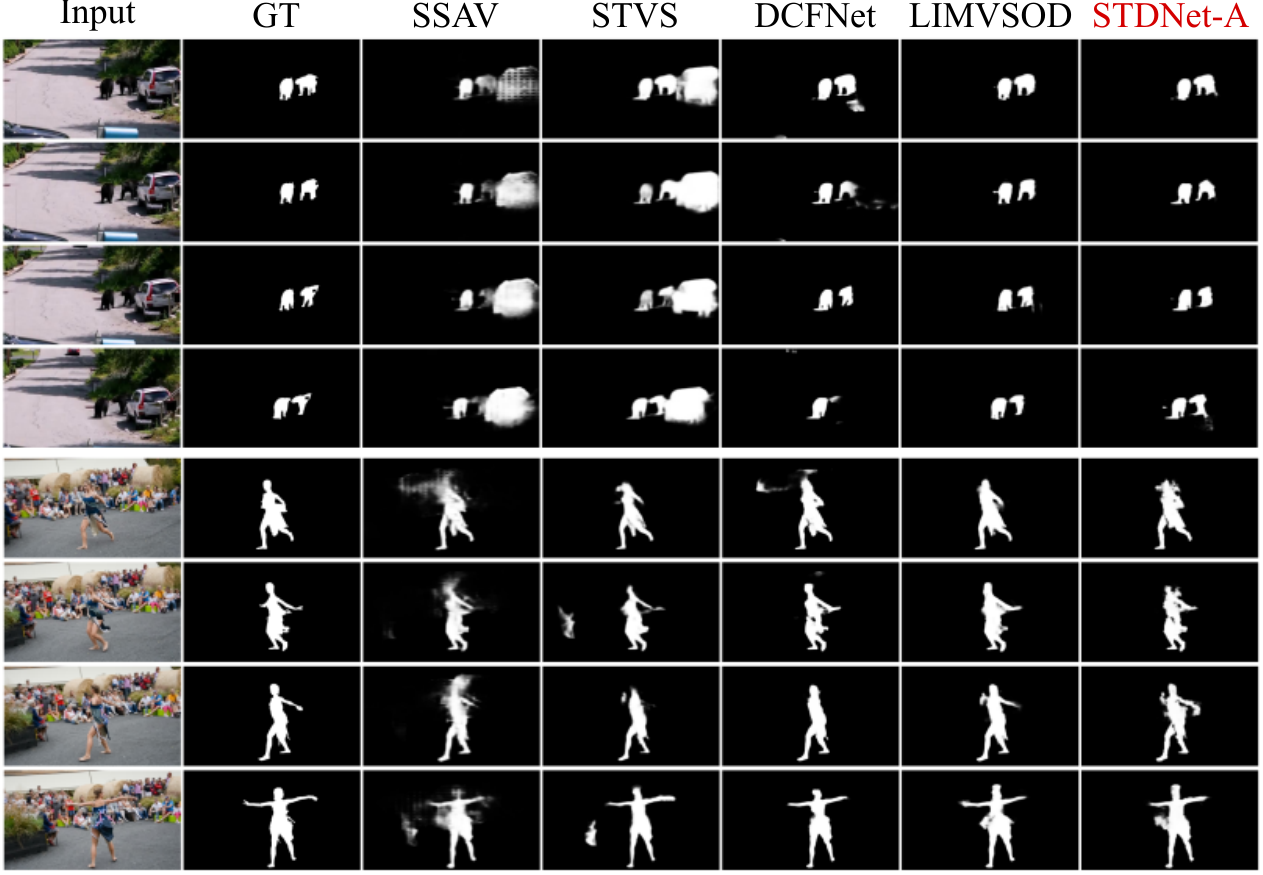}
    \caption{Qualitative comparison on VSOD. The first two columns are the input images and ground truth images respectively. Other columns contain saliency maps from different models.
    }
    \label{fig:quality_vsod}
\end{figure}

To investigate the necessity of ImageNet pretraining on ISOD models, we also conduct a comparison of models with pre-trained backbones, including both the large-backbone based ones~\cite{zhang2017ucf,zhang2017amulet,wang2017srm,wang2018dgrl,liu2019poolnet,zhao2019egnet,zhuge2022icon,wu2022edn,liu2021vst} and the latest lightweight ones~\cite{liu2021samnet,liu2020hvpnet,wu2022edn}. \textcolor{black}{In this case, we adopt the original resolutions used for generating the saliency maps.} The quantitative results are present in \cref{tab:pretrain} and a qualitative comparison is shown in the right part of~\cref{fig:quality}. It is interesting that SDNet-A with ViT blocks shows a better accuracy gain than SDNet in this case, which suggests that the global attention module needs more training data to compensate for its lack of inductive bias. 
The fact that SDNet does not have an improvement is in line with the conclusion in~\cite{cheng2021csnet} that ImageNet pretraining is not always needed for training lightweight ISOD models. However, we may update this conclusion since our ViT model needs it.

Overall, although the large models achieve better predictive performance, they suffer from prohibitive large model size and low inference speed. In contrast, SDNet-A achieves competitive accuracy with the fastest speed and low memory consumption.

\vspace{0.3em}
\noindent \textbf{Comparison on VSOD.}\quad
Based on the observations on ISOD, we train our STDNet and STDNet-A without and with ImageNet pretraining respectively. The quantitative and qualitative results are illustrated in~\cref{tab:vsod} and~\cref{fig:quality_vsod}. 

Again, both STDNet and STDNet-A employ extremely lightweight architectures with less than 1M parameters, making it easy to deploy on resource-limited edge devices. STDNet-A runs at 480 FPS and 150 FPS on the RTX 2080 Ti and AGX Orin respectively, which is more than $4\times$ and $3\times$ faster than the STVS method~\cite{chen2021stvs} with comparable or even better accuracy. It should be noted that STVS also aimed for a real-time model. More surprisingly, tested on the NX device with a more strict resource limitation, STDNet and STDNet-A still achieve real-time latency at about 28 FPS, while STVS only runs at 6.6 FPS.  When compared with other methods, the runtime advantage of our model is even larger. Accuracy-wise, prediction metrics are either improved or not significantly reduced with our models comparing with prior lightweight counterparts. This demonstrates the effectiveness of our STDC-based modules and network architectures. 

\begin{table}[t!]
\captionsetup{labelfont={color=black},font={color=black}}
\caption{Adopting EfficientNet-B5 in the STDNet backbone moves the trade-off towards accuracy.}
\centering
\setlength{\tabcolsep}{0.004\linewidth}
\resizebox*{\linewidth}{!}{
{\color{black}
\begin{tabular}{lcccccc|ccc|ccc}
\toprule
Model & \multirow{2}{*}{\begin{tabular}{@{}c@{}}\#Params\\(M)\end{tabular}} & \multirow{2}{*}{\begin{tabular}{@{}c@{}}FLOPs\\(G)\end{tabular}} & \multirow{2}{*}{\begin{tabular}{@{}c@{}}FPS\\(2080 Ti)\end{tabular}} & \multicolumn{3}{c}{DAVSOD} & \multicolumn{3}{c}{VOS} & \multicolumn{3}{c}{DAVIS} \\
& & & & $S$ & $F$ & M & $S$ & $F$ & M & $S$ & $F$ & M \\
\midrule
LIMVSOD~\cite{chen2022limvsod} & - & -& 2.4 & .792 & .725 & .064 & .844 & .822 & .060 & .922 & .911 & .016 \\
STDNet-A & 0.99 & 4.37 & 482 & .755 & .663 & .087 & .852 & .799 & .065 & .884 & .858 & .024 \\
STDNet (Eff) & 5.49 & 4.92 & 280 & .768 & .681 & .078 & .866 & .815 & .051 & .893 & .867 & .020 \\
\bottomrule
\end{tabular}
}
}
\label{tab:efficientnet}
\end{table}

\begin{table*}[t!]
\caption{Ablation study on different architecture settings. Scale means the width mutiplier of the model to expand the channels in each layer. FPS is calculated on the AGX Orin device. We mark the results from SDNet \textbf{in bold} if it performs the best, and \underline{underlined} if the second best.}
\centering
\setlength{\tabcolsep}{0.008\linewidth}
\resizebox*{\linewidth}{!}{
\begin{tabular}{lclccc|cc|cc|cc|cc|cc}
\toprule
\multirow{2}{*}{Scale} & \multirow{2}{*}{\begin{tabular}{@{}c@{}}FPS\\(Orin)\end{tabular}} & \multirow{2}{*}{Model} & \multirow{2}{*}{\begin{tabular}{@{}c@{}}Input\\size\end{tabular}} & \multicolumn{2}{c}{ECSSD} & \multicolumn{2}{c}{PASCAL-S} & \multicolumn{2}{c}{DUT-O} & \multicolumn{2}{c}{HKU-IS} & \multicolumn{2}{c}{SOD} & \multicolumn{2}{c}{DUTS-TE} \\
& & & & $F_\beta^{m}\uparrow$ & MAE$\downarrow$ & $F_\beta^{m}\uparrow$ & MAE$\downarrow$ & $F_\beta^{m}\uparrow$ & MAE$\downarrow$ & $F_\beta^{m}\uparrow$ & MAE$\downarrow$ & $F_\beta^{m}\uparrow$ & MAE$\downarrow$ & $F_\beta^{m}\uparrow$ & MAE$\downarrow$ \\
\midrule
\multirow{8}{*}{$\times 1.0$} & \multirow{4}{*}{76} & Baseline & 240$^2$ & .884 & .065 & .780 & .100 & .701 & .080 & .864 & .061 & .743 & .135 & .748 & .076 \\
& & Baseline-Rep & 240$^2$ & .882 & .066 & .786 & .100 & .707 & .080 & .864 & .061 & .741 & .136 & .751 & .077 \\
& & PiDiNet~\cite{su2021pidinet} & 240$^2$ & .877 & .070 & .771 & .106 & .702 & .083 & .854 & .068 & .735 & .140 & .737 & .081 \\
& & SDNet & 240$^2$ & \textbf{.890} & \textbf{.064} & \textbf{.787} & \textbf{.099} & \textbf{.712} & \textbf{.080} & \textbf{.869} & \textbf{.060} & \textbf{.749} & \textbf{.134} & \textbf{.754} & \underline{.077} \\
\cmidrule(r){2-16}
& \multirow{4}{*}{46} & Baseline & 320$^2$ & .904 & .056 & .806 & .093 & .735 & .076 & .889 & .052 & .779 & .123 & .787 & .070 \\
& & Baseline-Rep & 320$^2$ & .900 & .057 & .807 & .094 & .736 & .076 & .888 & .052 & .776 & .126 & .786 & .070 \\
& & PiDiNet~\cite{su2021pidinet} & 320$^2$ & .898 & .060 & .793 & .101 & .728 & .080 & .881 & .057 & .767 & .130 & .774 & .075 \\
& & SDNet & 320$^2$ & \underline{.903} & \underline{.057} & \textbf{.809} & \textbf{.093} & \textbf{.737} & \underline{.077} & \textbf{.891} & \textbf{.052} & \textbf{.780} & \textbf{.122} & \underline{.786} & \textbf{.070} \\
\midrule
\multirow{4}{*}{$\times 0.75$} & \multirow{4}{*}{54} & Baseline & 320$^2$ & .887 & .067 & .789 & .105 & .714 & .087 & .873 & .060 & .758 & .134 & .762 & .081 \\
& & Baseline-Rep & 320$^2$ & .896 & .061 & .803 & .096 & .728 & .079 & .880 & .057 & .763 & .128 & .770 & .075 \\
& & PiDiNet~\cite{su2021pidinet} & 320$^2$ & .877 & .075 & .770 & .119 & .704 & .098 & .853 & .075 & .748 & .142 & .735 & .096 \\
& & SDNet & 320$^2$ & \textbf{.897} & \underline{.063} & \underline{.796} & \underline{.101} & \underline{.726} & \underline{.084} & \textbf{.883} & \textbf{.057} & \textbf{.772} & \textbf{.128} & \textbf{.771} & \underline{.078} \\
\bottomrule
\end{tabular}
}
\label{tab:ablation-dcr}
\end{table*}

\textcolor{black}{
Since our models have already achieved remarkable inference speed, it gives space to move the trade-off towards the accuracy side while still maintaining the efficiency advantage. For example, adopting EfficientNet-B5~\cite{tan2019efficientnet} in the backbone of STDNet combing with our STDC-based spatiotemporal modules in the decoder, the accuracy gap between STDNet and the state-of-the-art accurate models can be reduced to some extent, and the running speed is nearly halved but still competitive (~\cref{tab:efficientnet}).
}

\subsection{Model analysis}
\label{sec:ablation}
In this part, ablation studies are presented to investigate individual components of our designs and models. 
The image and video models are trained with 60 epoches for ISOD and VSOD respectively with other configurations kept the same as above. As a normal routine, we choose the best performed models as our final models (\ie SDNet for ISOD and STDNet-A for VSOD), and observe the changes in performance by alternate the design process. Precisely, we are concerned about: 1) How does DCR enhance the original PiDiNet backbone on ISOD? 2) Is the extraction of high-order feature contrasts by PDCs necessary on ISOD? 3) How does the design of STDNet impacts the temporal consistency and final accuracy when it comes to videos on VSOD? 5) Other factors that affect the model performance.

\vspace{0.3em}
\noindent \textbf{Effectiveness of DCR.}\quad
Recalling that the DCR transforms the multi-branch layer in the backbone to a single standard convolutional layer, concerns about the necessity of the additional PDC operators during training are raised. To validate the positive roles played by PDCs and DCR, we construct the following backbone variants:
\begin{itemize}[leftmargin=*]
    \item \textbf{Baseline}: using a PiDiNet-like backbone, where each layer has a single branch but only using the standard convolutions.
    \item \textbf{Baseline-Rep}: using a SDNet-like backbone, where each layer has the same number of branches during training as in SDNet but only using the standard convolutions. After training, each layer is reparameterized into a single-branch version.
    \item \textbf{PiDiNet}: using the original PiDiNet backbone, \ie each layer has a single branch using one of the following operators sequentially: CPDC, APDC, RPDC, and the standard convolution.
    \item \textbf{SDNet}: the proposed backbone.
\end{itemize}
We compare them in two different scales and input sizes. As shown in~\cref{tab:ablation-dcr}, ``Baseline-Rep'' fails to provide a performance gain with respect to ``Baseline'' in several cases. This suggests that the reparameterization has a limited effect when applied solely to standard convolutions. Meanwhile, SDNet achieves the best performance in most cases, demonstrating that employing PDCs is beneficial for SOD with explicit feature contrast measuring. 

\begin{figure}[t!]
    \centering
    \includegraphics[width=\linewidth]{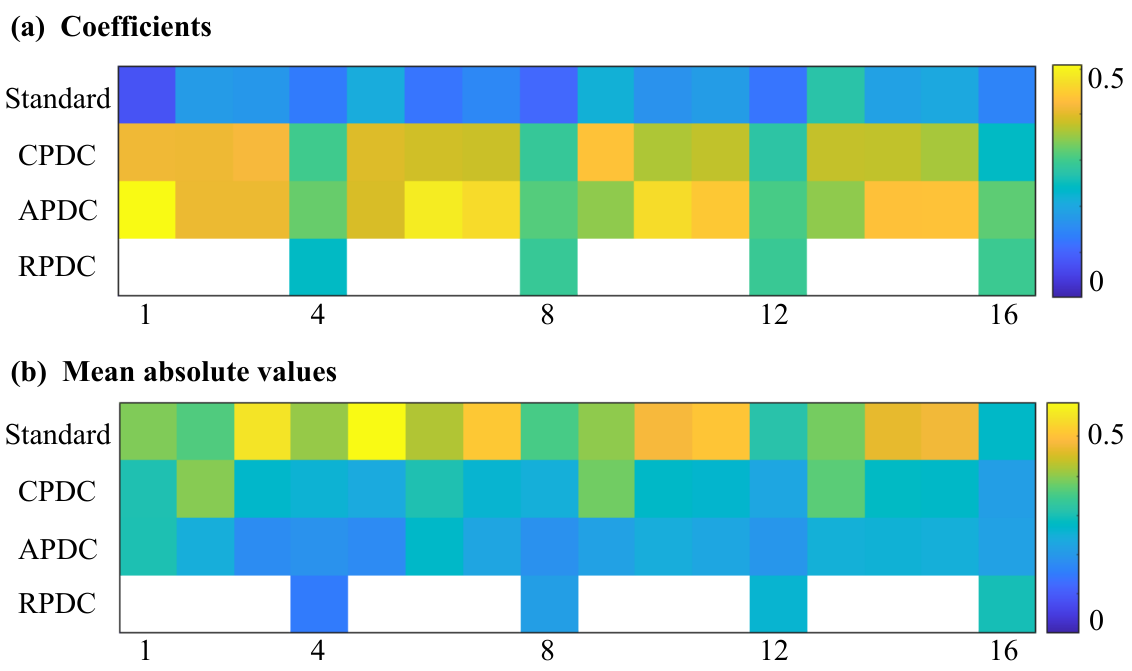}
    \caption{For both figures, each column represents a certain layer in SDNet, where the corresponding values of different types of convolution are shown, noting that RPDC is only adopted in layer 4, 8, 12, and 16: (a) The learned coefficients (in our experiments, we use softmax function to constrain the sum of coefficients in each layer to 1). (b) The averaged absolute values from the output of each individual convolutional branches before branch fusion (we normalized them to [0, 1] by dividing them with the sum for each layer). Statistics are based on ECSSD dataset
    }
    \label{fig:coefficients}
\end{figure}

\vspace{0.3em}
\noindent \textbf{Contributions of individual operators.}\quad
We can track the contributions of individual branches through the corresponding coefficients $\{\alpha_i\}$ and output values of these branches before fusion to get some insights. As shown in~\cref{fig:coefficients}, the coefficients of PDC operators are larger, while the final output values of PDC branches are smaller when compared with the corresponding values of standard convolution. Since the pixel differences usually have lower values than pixel intensities due to the fact that pixels in local regions share similar values, the coefficients of PDC operators should be larger to make their outputs comparable to those of standard convolution. Meanwhile, the lower output values of PDC operators indicate that SOD models rely more on the low-frequency features from standard convolution, as most of the salient regions are composed of low-frequency signals (\eg the large inner part of objects).

\begin{table}[t!]
\caption{Ablation study on temporal modules, frame number, and padding methods (replicate padding by default). In STDM, ``sc'', ``cd'', and ``ad'' mean using standard convolution, CSTDC, and ASTDC, respectively. All the STDM modules share the same running latency thanks to our DCR strategy. 
Our final model is marked in \textbf{bold}. We mark its results \textbf{in bold} if it performs the best, and \underline{underlined} if the second best. }
\centering
\setlength{\tabcolsep}{0.008\linewidth}
\resizebox*{\linewidth}{!}{
\begin{tabular}{lcccc|ccc}
\toprule
\multirow{2}{*}{\begin{tabular}{@{}c@{}}Temporal\\module\end{tabular}}  & \#frames & \multicolumn{3}{c}{DAVSOD (2019)} & \multicolumn{3}{c}{VOS (2018)} \\
& & $S_\lambda\uparrow$ & $F_\beta^{m}\uparrow$ & MAE$\downarrow$ & $S_\lambda\uparrow$ & $F_\beta^{m}\uparrow$ & MAE$\downarrow$ \\
\midrule
N/A & N/A & .705 & .604 & .092 & .835 & .783 & .065 \\
\midrule
STDM (sc) & 8 & .724 & .616 & .095 & .839 & .790 & .067 \\
STDM (cd) & 8 & .739 & .641 & .091 & .840 & .792 & .061 \\
STDM (ad) & 8 & .747 & .653 & {.081} & .848 & .796 & .063 \\
\midrule
STDM (sc+cd+ad) & 2 & .721 & .619 & .093 & .842 & .791 & .069 \\
STDM (sc+cd+ad) & 4 & .751 & {.669} & {.082} & .844 & .792 & .065 \\
\textbf{STDM (sc+cd+ad)} & {8} & \textbf{.755} & \underline{.663} & .087 & \textbf{.852} & \textbf{.799} & .065 \\
\textcolor{black}{STDM (cv+cd+ad)} & \textcolor{black}{16} & \textcolor{black}{.720} & \textcolor{black}{.614} & \textcolor{black}{.097} & \textcolor{black}{.849} & \textcolor{black}{.799} & \textcolor{black}{.061}
\\
\multirow{2}{*}{\begin{tabular}{@{}c@{}}STDM (sc+cd+ad)\\w/ zero padding\end{tabular}} & \multirow{2}{*}{8} & \multirow{2}{*}{.731} & \multirow{2}{*}{.627} & \multirow{2}{*}{.092} & \multirow{2}{*}{.848} & \multirow{2}{*}{.793} & \multirow{2}{*}{.063} \\
& & & & & & & \\
\midrule
Tempral attentions & 8 & .732 & .635 & .090 & .852 & .798 & .060 \\
\bottomrule
\end{tabular}
}
\label{tab:vsod-ablation-testset}
\end{table}

\begin{figure}[t!]
    \centering
    \includegraphics[width=\linewidth]{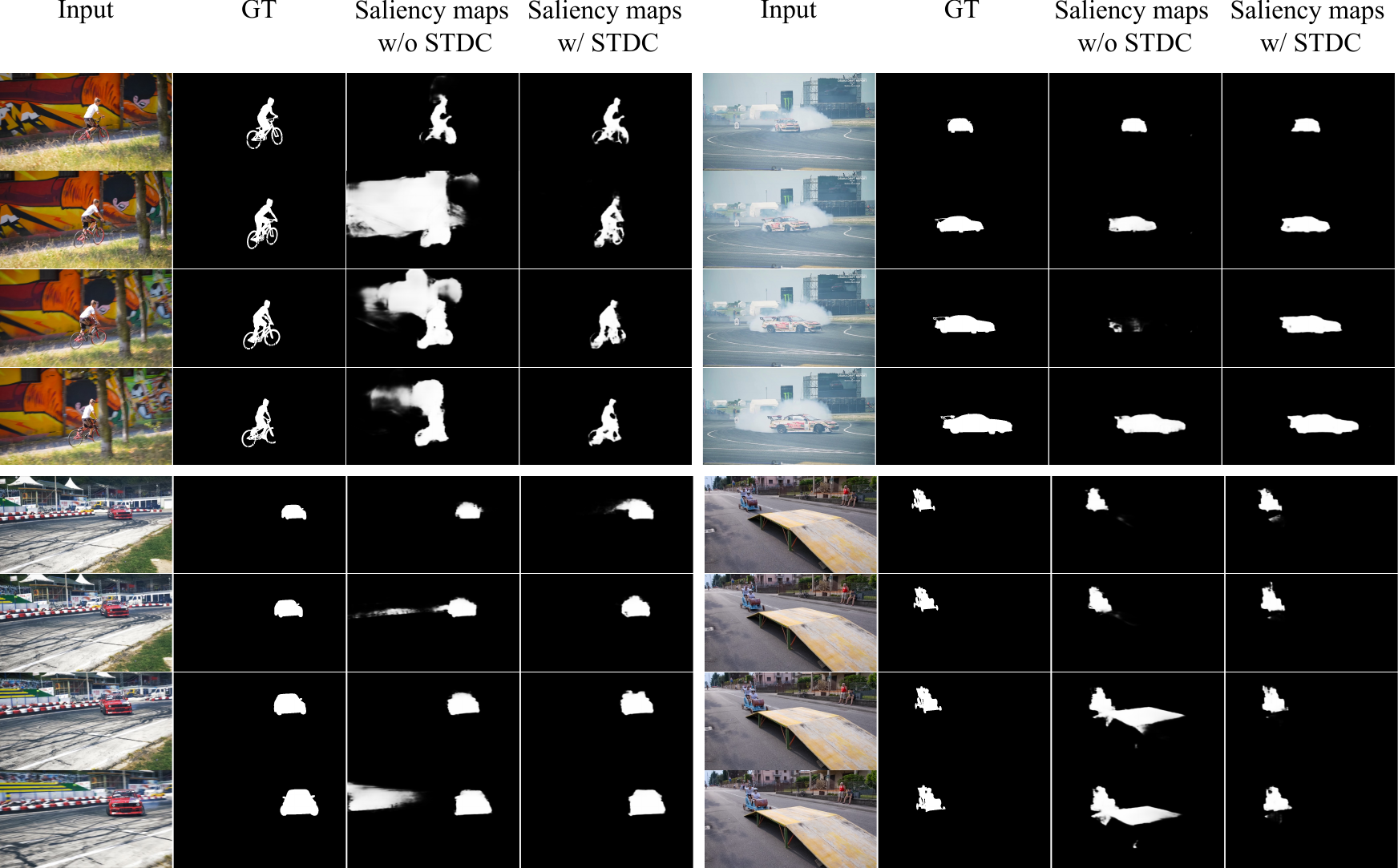}
    \caption{Predicted saliency maps w/ and w/o temporal modules. It can be seen that the temporal inconsistency of model without temporal modules leads to weak prediction robustness though the input images look similar. It is remedied by our STDM, which leads to better consistency and qualities of final predictions.
    }
    \label{fig:temporal-consistency}
\end{figure}

\vspace{0.3em}
\noindent \textbf{Temporal consistency.}\quad
STDM is the only spatiotemporal module in the proposed STDNet which considers motion information via STDC. Recalling the STDNet structure in~\cref{fig:stdnet}, we have used CDCM~\cite{pidinet-pami} and STDM in parallel to extract spatial and spatiotemporal features respectively. Thereby, we replace our STDM with CDCM to check the impact of only utilizing spatial modules (denoted as ``N/A'' in~\cref{tab:vsod-ablation-testset}). The metric results degrade by a large margin on both DAVSOD and VOS, demonstrating the importance of temporal cues. We can see from~\cref{fig:temporal-consistency} that STDM effectively enhances the temporal consistency of video frames, resulting in better saliency maps.

\begin{figure}[t!]
    \centering
    \includegraphics[width=\linewidth]{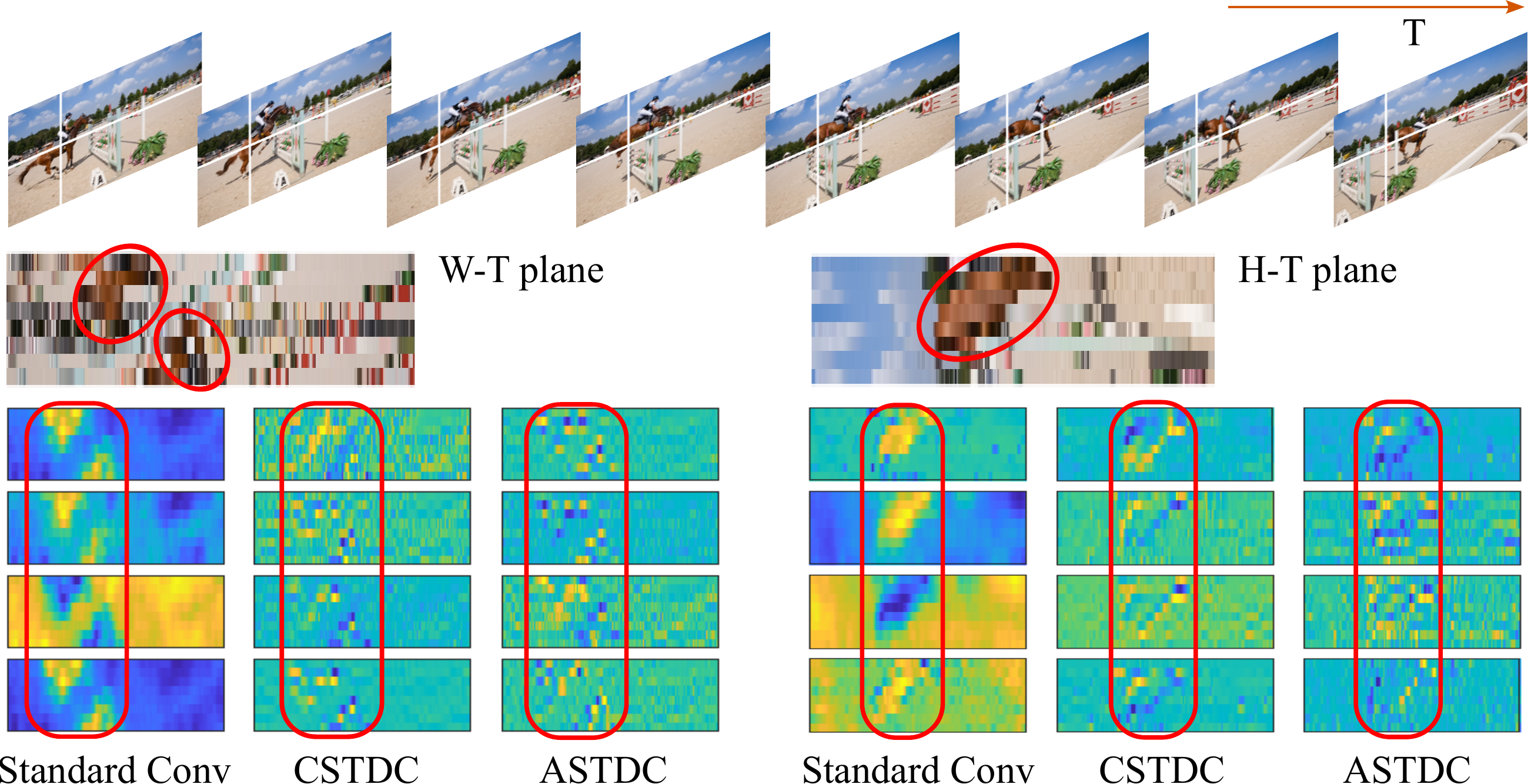}
    \caption{Visualization of spatiotemporal feature maps by different convolutional operators. The standard convolution and STDC capture spatiotemporal features in a complementary way, where the standard convolution generates features maps with mostly zeroth-order intensities, while STDC focuses more on the higher-order spatiotemporal contrasts.
    }
    \label{fig:tdc_features}
\end{figure}

\vspace{0.3em}
\noindent \textbf{Effectiveness of STDC.}\quad
To validate the effectiveness of STDC, we build multiple STDM variants that incorporate different convolution types. DCR enables us to add an arbitrary number of convolution types in STDM without affecting the runtime efficiency. When only one type is adopted, as listed in the third part of~\cref{tab:vsod-ablation-testset}, the standard convolution performs worse than either CSTDC or ASTDC, whereas ASTDC alone performs the best (noting that the standard convolution-based STDM still captures spatiotemporal information, since the execution space of convolution is W-T and H-T planes). This implies that the high-order spatiotemporal contrasts (in the form of pixel differences) are more important in encoding spatiotemporal features than the zeroth-order intensity information commonly encoded by standard convolutions. When combining both STDCs and standard convolution (denoted by ``sc+cd+ad''), the model gives the best performance. A visualization result is also shown in~\cref{fig:tdc_features} for the standard convolution and STDC respectively. 

\vspace{0.3em}
\noindent \textbf{Number of frames; padding method.}\quad
The fourth part of~\cref{tab:vsod-ablation-testset} shows that more frames in the input generally lead to higher accuracy. 
\textcolor{black}{
However, the performance saturates at 8 and declines with more frame numbers. The phenomenon might be caused by the limited receptive field along the temporal dimension of the STDM in the side structure. We found that adding more STDMs (to enlarge the receptive field) might affect the efficiency and gives no accuracy gain.} 
Based on that, we set the number of frames to 8 in our final model. We also observe that changing replicate padding to zero padding degrades performance, due to the small temporal dimension. 

\begin{figure}[t!]
    \centering
    \includegraphics[width=\linewidth]{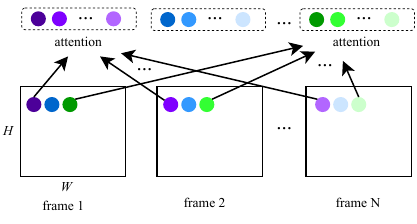}
    \caption{Temporal attentions. Features of each frame is regarded as a token. Attentions are conducted on each pixel location separately.
    }
    \label{fig:temporal-attention}
\end{figure}

\vspace{0.3em}
\noindent \textbf{Exploration on temporal attentions.}\quad
While we use STDCs to encode temporal cues, it would be interesting to compare it with attention modules based on ViTs. To capture global feature correlations along the temporal dimension, it is straightforward to regard each video frame as a token and design a lightweight MobileViT-like attention temporal module. Such a module is designed with temporal attentions, as illustrated in \cref{fig:temporal-attention}. Similar to~\cite{mehta2022mobilevitv2}, we reshape the input features $\pmb{F}\in\mathbb{R}^{C\times T\times H\times W}$ ($C$, $T$, $H$, $W$ represents number of feature channels, number of frames, height, and width, respectively) to $\pmb{F}'\in\mathbb{R}^{HW\times T\times C}$. By setting $HW$ as the batch size, $T$ as the number of tokens, and $C$ as the number of channels in token features, the normal transformer block~\cite{vaswani2017attentionallyouneed} can then be constructed. 

The attention-based temporal module serves as a strong baseline with its capability to extract local and global correlations between frames. As can be seen in \cref{tab:vsod-ablation-testset}, attention module beats the standard convolution-based STDM by a considerable margin, \ie 0.724 \vs 0.732 of $S_\lambda$ on DAVSOD and 0.839 \vs 0.852 of $S_\lambda$ on VOS for standard convolutions \vs attentions. However, attention modules suffer from its limitation on using zeroth-order intensities. The incorporation of high-order spatiotemporal contrast cues by STDC gives more benefits (\eg $S_\lambda$ is improved to 0.755 on DAVSOD), even though it only utilizes convolutional operators. 

\section{Conclusion}
\label{sec:conclusion}

This work addresses the critical task of Salient Object Detection (SOD), crucial in various computer vision applications. 
Our contribution lies in proposing a novel approach that combines classical heuristic insights with the capabilities of CNNs, aiming to achieve a delicate balance between speed and accuracy. By focusing on contrast cues and utilizing Pixel Difference Convolutions (PDCs), we develop lightweight and efficient SOD models. The integration of the proposed Difference Convolution Reparameterization (DCR) strategy further streamlines our models, ensuring both effectiveness and efficiency. In addition to SOD for single images, we also extend our methodology to videos through novel SpatioTemporal Difference Convolutions (STDC), with which spatiotemporal contrasts are better encoded. 

Benchmarking our models on consumer-grade GPUs and embedded systems, we have demonstrated noteworthy improvements in efficiency-accuracy trade-offs compared to existing lightweight methods. Notably, our model exhibits exceptional real-time performance on both Image SOD (ISOD) and Video SOD (VSOD), outperforming competitors in terms of running speed and prediction results.

In essence, this work not only contributes novel models for efficient SOD but also underscores the importance of marrying classical wisdom with modern deep learning techniques. As the demand for real-time processing intensifies, our proposed approach serves as a promising step towards achieving the delicate equilibrium between efficiency and accuracy in the field of Salient Object Detection. 

\textcolor{black}{
Lastly, as ``abstract operators'', PDC and STDC can be instantiated with flexible forms (\eg CPDC, APDC, CSTDC, ASTDC). These different and multi-functional operators are a free lunch due to our DCR strategy. Not limited to SOD, many other applications may benefit from it. We hope more future works can be inspired from this paper, for instance, tracking, remote sensing, and satellite imagery applications, where contrast cues are important components of visual features.
}

\footnotesize
\bibliographystyle{IEEEtran}
\bibliography{IEEEabrv,main_arxiv}

\end{document}